\documentclass{bmvc2k}

\title{Post-Training VLMs for Video Mistake Detection}

\addauthor{Federico Spurio*}{fspurio@iai.uni-bonn.de}{1,2}
\addauthor{Olga Zatsarynna*}{zatsarynna@iai.uni-bonn.de}{1,2}
\addauthor{Lars Doorenbos*}{doorenbos@iai.uni-bonn.de}{1,2}
\addauthor{Emad Bahrami}{bahrami@iai.uni-bonn.de}{1}
\addauthor{Gianpiero Francesca}{gianpiero.francesca@gmail.com}{3}
\addauthor{Juergen Gall}{gall@iai.uni-bonn.de}{1,2}

\addinstitution{
 University of Bonn\\
 Bonn, Germany
}
\addinstitution{
 Lamarr Institute for Machine Learning and Artificial Intelligence\\
 Bonn, Germany
}
\addinstitution{
 Toyota Motor Europe\\
 Belgium
}

\runninghead{Spurio et al.}{Post-Training VLMs for Video Mistake Detection}

\def\ie{\emph{i.e}\bmvaOneDot}

\def\eg{\emph{e.g}\bmvaOneDot}

\newif\ifdraft
\drafttrue

\definecolor{orange}{rgb}{1,0.5,0}
\definecolor{gr}{rgb}{0,0.65,0}
\definecolor{mygray}{gray}{0.95}

\ifdraft
 \newcommand{\RS}[1]{{\color{red}{\bf RS: #1}}}
 
 \newcommand{\PMN}[1]{{\color{orange}{\bf PMN: #1}}}

\else
 \renewcommand{\sout}[1]{}
 \newcommand{\RS}[1]{{\color{red}{}}}
 
 \newcommand{\PMN}[1]{{\color{red}{}}}
 
\fi

\newcommand{\corrans}{a}
\newcommand{\oppans}{\tilde{a}}
\newcommand{\gtans}{\upalpha}
\begin{document}

\maketitle

\begin{abstract}
\noindent
Human mistakes are inevitable when following instructions, yet they can lead to severe consequences. As such, there has been an increased interest in developing methods for detecting mistakes in videos, with current methods mostly focusing on closed-set protocols. While successful in controlled settings, the closed-set assumption limits their wider applicability, as any changes to the task require collecting new data and re-training models. 
Instead, we argue that mistake detection methods should learn the general concept of a mistake, rather than overfitting to step-specific details.
To reflect this, we introduce the Mistake Detection Video Question Answering (MD-VQA) protocol and accompanying benchmark. MD-VQA tests whether methods can discern if a step was executed correctly with respect to its description, for both seen and unseen actions. 
To address this important challenge, we propose the first video-language-model post-training technique for mistake detection.
Our method uses a tailored reward function to encourage the model to identify discrepancies between an instruction and the corresponding video. 
Extensive evaluations demonstrate that this approach outperforms zero-shot, supervised fine-tuning, and post-training baselines. Notably, our method generalizes especially well to unseen procedures, for instance, with an improvement of up to $11.6\%$ over the best-performing baseline on EP-VQA, paving the way toward general mistake detection. We release our code and benchmark at \url{https://github.com/FedeSpu/mstk}.

\end{abstract}

%-------------------------------------------------------------------------
\section{Introduction}
\label{sec:intro}

Human mistakes are as costly as they are common. Across a wide range of domains and activities, such as surgeries, driving, and assembly lines, these errors can lead to significant losses in efficiency, productivity, and resources (\eg,~\cite{suliburk2019analysis}). Generally, mistakes do not stem from a lack of expertise or motivation on the part of the individuals involved, but rather are inherent to any processes involving humans. As a result, since the occurrence of mistakes is unavoidable, they should be addressed appropriately by designated systems that understand when errors occur. 

For this task, artificial intelligence has substantial potential, as also evidenced by the growing number of works addressing video mistake understanding. Given the highly multifaceted nature of this problem, existing approaches adopt diverse formulations. Prior works have explored mistake understanding from the perspectives of online error detection~\cite{flaborea2024prego, seminara2024differentiable, patsch2025ICCV, mazzamuto2024eyes}, temporal action and error segmentation~\cite{lee2024egoper, lee2025error, guo2025procedural, huang2025modeling}, per-clip error classification~\cite{HoloAssist2023}, and zero-shot long-form mistake detection~\cite{peddi2024captaincook4d}.
Many of these methods demonstrate strong performance on their respective protocols; however, one key aspect of mistake understanding has been largely underexplored - \textbf{effective mistake detection in unseen scenarios}. Instead, most of the previously developed models operate under a closed-set assumption, where the action classes for which mistakes must be detected are assumed to be shared between the training and inference stages. For example, \cite{flaborea2024prego} focuses on online detection of procedural errors, such as executing actions in the wrong order or skipping required steps. While these approaches do not assume that the training data contains errors, they assume that all actions are part of the training set. Therefore, such methods can not be applied to unseen scenarios with actions that are not part of the training data.
This significantly limits the adaptability and scalability of such approaches.

\begin{figure*}[t!]
    \centering
    \includegraphics[width=\linewidth]{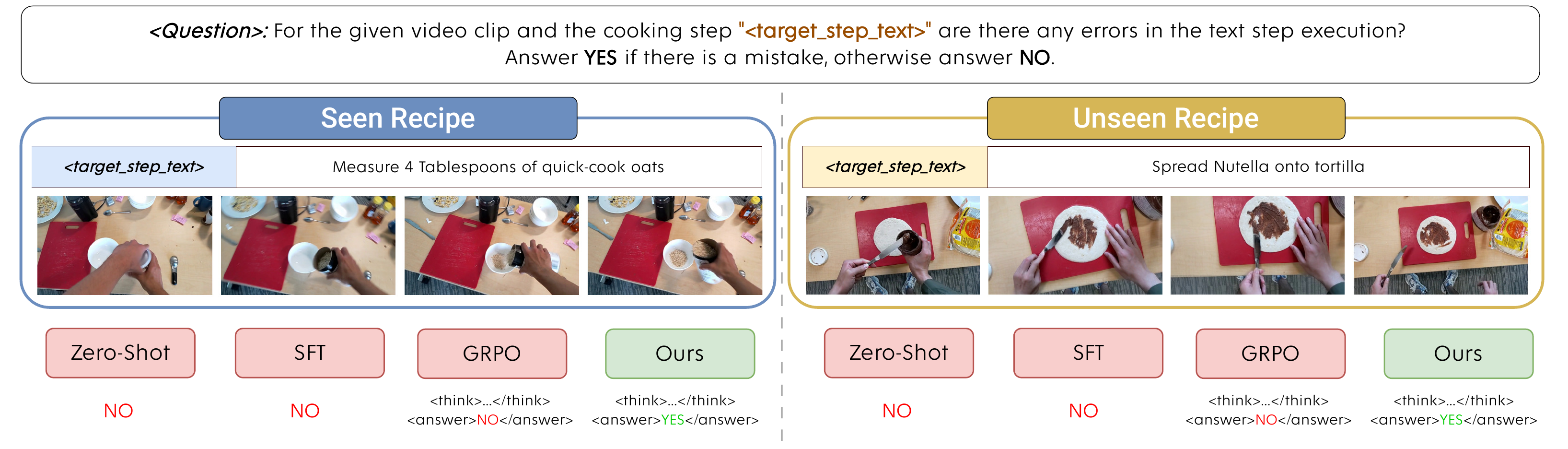}
    \caption{\textbf{Mistake detection in seen and unseen domains.} 
    Our benchmark evaluates models not only on recipes present in the training data (left), but also tests their ability to generalize the abstract concept of ``mistake'' to novel tasks (right). 
    While standard training strategies (Zero-Shot, SFT, and GRPO) fail to detect these subtle errors or struggle to generalize, our proposed post-training approach successfully grounds its reasoning to accurately identify mistakes across both scenarios. Examples from EP-VQA.}
    \label{fig:teaser}
    \vspace{-0.5cm}
\end{figure*}

So far, only one work~\cite{peddi2024captaincook4d} has addressed open-set scenarios by leveraging VLMs in a zero-shot fashion. It introduced a Video Question Answering (VQA) style setup for mistake understanding in a zero-shot, long-form manner. However, such a formulation leads to an overly difficult problem setting, as shown by the low performance of zero-shot VLMs on the proposed task. The difficulty arises from combining two individually challenging tasks: zero-shot temporal grounding of short segments in very long videos (over 20 minutes) and zero-shot detection of fine-grained mistakes.
Given the discouraging results in~\cite{peddi2024captaincook4d}, we introduce a new protocol. While we also formulate mistake detection as a VQA task, we focus on detecting errors within video segments corresponding to a single instruction step, rather than using a full-video long-horizon approach. More specifically, given a procedural step and its corresponding video clip, the task is to determine whether the execution shown in the clip matches the provided textual instruction, as illustrated in Fig.~\ref{fig:teaser}. To establish a benchmark for this new setting, we adapt and unify the CaptainCook4D~\cite{peddi2024captaincook4d} and EgoPER~\cite{lee2024egoper} datasets to our protocol. As a result, our benchmark includes a diverse and challenging set of questions spanning various complex instruction steps and mistake types.

We show that, also on our benchmark, zero-shot VLM performance (as proposed in \cite{peddi2024captaincook4d}) remains poor, since detecting mistakes in fine-grained instructional steps requires complex multi-step reasoning.
Therefore, inspired by the success of post-training in tasks requiring high-level reasoning, we propose a reinforcement–learning–based post-training algorithm designed for mistake detection as our second contribution. Our method uses a novel reward function that samples alternative video clips for each instruction–video pair to better identify mismatches between textual and visual information. We compare our approach against a variety of baselines and demonstrate that our post-training method achieves state-of-the-art performance, especially on unseen scenarios. For instance, we improve the F1-score by $5.0$ over the next-best baseline, an improvement of 11.6\%.

In short, our main contributions can be summarized as follows:
\begin{itemize}
    \item We propose a novel Mistake Detection VQA protocol, which we term MD-VQA, for both seen and unseen instructional steps. 
    \item We present the first VLM post-training technique for effectively addressing MD-VQA.
\end{itemize}

%-------------------------------------------------------------------------
\section{Related Work}
\label{sec:related-work}

\subsection{Mistake Understanding}
\label{sec:rw_mistake}
Before mistake understanding emerged as a distinct research direction, several related lines of work were commonly addressed, such as video anomaly detection~\cite{gong2019memorizing, Sabokrou2017DeepCascadeC3, park2020learning, liu2018future, lee2024egoper, Sultani2018RealWorldAD, Wu2020NotOL, tian2021weakly} and unintentional action prediction~\cite{epstein2020oops, zatsarynna2022gcpr, duka2022leveraging, epstein2021learning}.
In recent years, research has increasingly focused on post-hoc detection and recognition of mistakes, for example, in assembly~\cite{ding2025mistakeassembly} and physical manipulation~\cite{HoloAssist2023} tasks.
Besides post-hoc detection, other lines of work address online mistake detection~\cite{flaborea2024prego, seminara2024differentiable, mazzamuto2024eyes, patsch2025ICCV}, or focus on error detection and/or recognition based on temporal action segmentation (TAS)~\cite{lee2024egoper, lee2025error, guo2025procedural, huang2025modeling}. 
Most recently, there has been an increase in the use of VLMs for mistake detection, motivated by their success in other video-related tasks. For instance, \cite{peddi2024captaincook4d} introduced a zero-shot approach that evaluates the alignment between long-form video clips and textual recipe steps, whereas other methods use zero-shot~\cite{lee2025error} or fine-tuned VLMs~\cite{patsch2025ICCV} to recognize and explain mistakes.
The above approaches demonstrate strong results under their respective formulations; however, these methods come with a crucial limitation: most existing approaches~\cite{lee2024egoper, lee2025error, guo2025procedural, huang2025modeling, seminara2024differentiable, patsch2025ICCV, HoloAssist2023} are intrinsically closed-set and cannot predict errors in previously unseen action steps. Although relying on zero-shot VLM capabilities~\cite{peddi2024captaincook4d} can mitigate this issue, it is often insufficient for strong performance on unseen actions.
In contrast, we introduce a new protocol explicitly focused on general mistake detection, where models need to identify mistakes in video segments for both seen and unseen instruction steps.

\subsection{Post-Training VLMs}
Reinforcement Learning (RL) is now the primary approach for post-training Vision-Language Models (VLMs). In this context, methods such as proximal policy optimization~\cite{schulman2017proximal} (PPO) enable better alignment with human preferences while retaining the reasoning capabilities of the original model. However, these methods can be inefficient due to reliance on an additional critic model. In contrast, Group Relative Policy Optimization (GRPO) \cite{shao2024deepseekmath} offers a computationally efficient alternative by eliminating the critic network and computing advantages relative to group mean rewards. 
Recent work has successfully applied GRPO and other RL post-training methods to video understanding tasks, such as Video-R1 \cite{feng2025videor1}, DeepVideo-R1~\cite{li2025deepvideor1} or VideoAuto-R1 \cite{liu2026videoauto} to improve video reasoning. For instance, VideoChat-R1 \cite{li2025videochatr1} enhances spatio-temporal perception through reinforcement fine-tuning, ArrowRL \cite{xue2025arrowrl} focuses on temporal reasoning capabilities, and EgoThinker~\cite{pei2025egothinkerunveilingegocentricreasoning} utilizes GRPO-based post-training to improve egocentric video understanding.
These approaches highlight GRPO's effectiveness in multi-modal settings, where traditional methods are computationally prohibitive.
Our method builds upon this line of work and introduces the first post-training method for mistake detection.

%-------------------------------------------------------------------------

%-------------------------------------------------------------------------
\section{MD-VQA: Video Question Answering for Mistake Detection}
\label{sec:bench}

In this section, we present our mistake detection protocol. Unlike most prior formulations, which treat mistake understanding as a closed-set task where the actions or instructional steps are shared between the training and test set, we focus on detecting mistakes in unconstrained instructional steps, enabling more adaptable and generalizable models. 
In contrast to \cite{peddi2024captaincook4d}, which focuses on a long-form setup that conflates temporal grounding and mistake detection, leading to poor performance across methods, we focus on the detection of fine-grained mistakes for a given instruction and short video clip.
Therefore, we set up a new evaluation benchmark to systematically test the performance of various VLM approaches on the proposed MD-VQA task. We also move beyond evaluating only training-free methods, guided by the poor zero-shot results even in our problem setting.

\subsection{MD-VQA Task}
\label{sec:bench_task}

\begin{figure}[tbp]
    \centering
    \includegraphics[width=\linewidth]{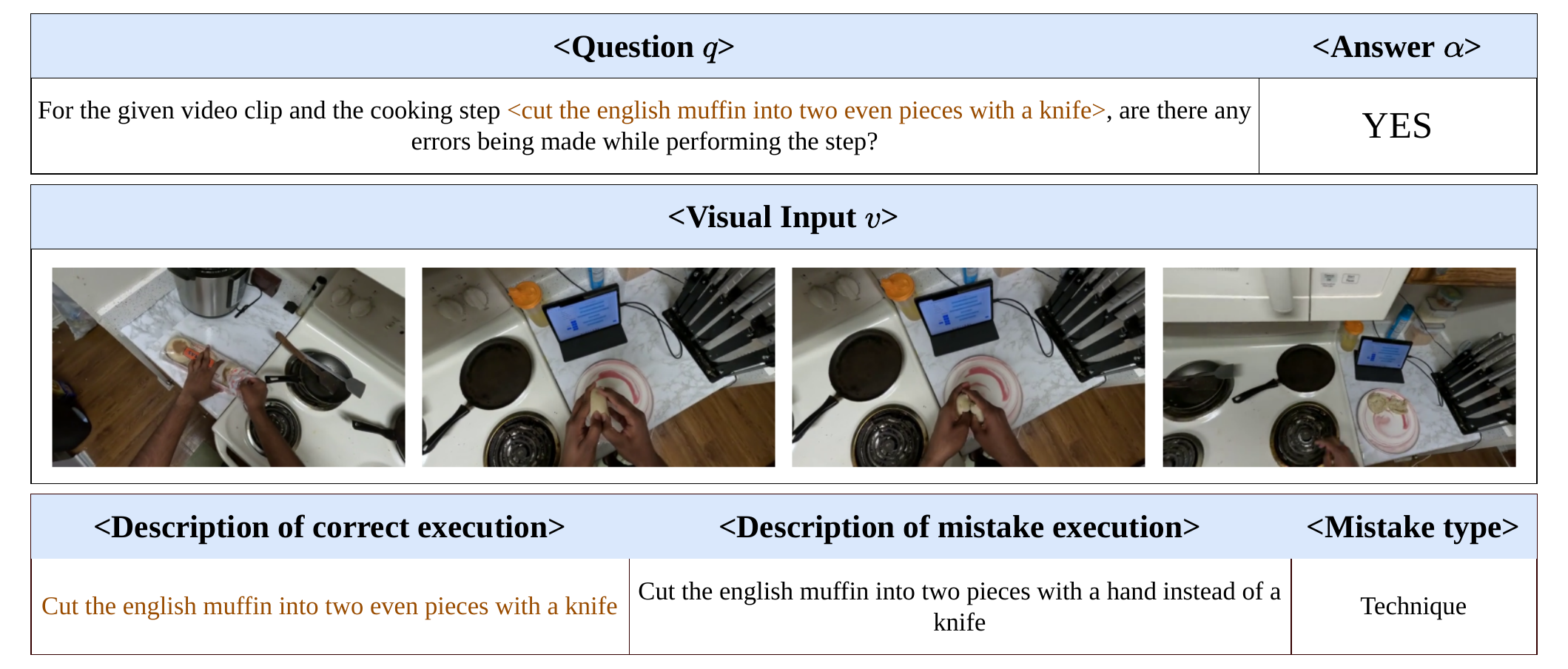}
    \caption{{\bf Example from CC-VQA Test Seen split.} Question $q$ asks whether the cooking step \textit{``cut the english muffin into two even pieces with a knife''} contains any mistakes in the video $v$. The ground-truth answer $\alpha_i$ is ``YES'' as the muffin is shredded by hand rather than cut with a knife.}
    \label{fig:benchmark}
    \vspace{-0.3cm}
\end{figure}

To assess whether VLMs can capture nuances across different instructional steps, we formulate mistake detection as a video question-answering task. Formally, given a video $v$ showing either a correct or faulty execution of an instructional step $t$, we construct a question $q$ that inquires whether the step $t$ was performed correctly or contains a mistake. We show an example of such a question in Fig.~\ref{fig:benchmark}. In our work, we focus on binary questions, where the VLM must determine the presence of an error. The ground-truth answers for such questions are $\gtans=\text{\textbf{yes}}$ if there is a mismatch between the instruction and its execution, or $\gtans=\text{\textbf{no}}$ otherwise.
In this way, each MD-VQA example can be represented as a tuple $m=\{v, q, \gtans\}$. 
We construct four paraphrased question variants for each $q$ that differ in wording.
For training, methods can use all four variants to improve linguistic diversity and reduce prompt overfitting. During inference, a single formulation of $q$ is used. The exact question templates are provided in the supplementary material.

It is important to note that we do not assume that the steps \(t_i\) come from a known pre-fixed set. Models are therefore required to recognize mistakes in both familiar and unseen instructional steps. Achieving strong performance in this setting thus demands that models learn a generalizable notion of what constitutes a mistake. This renders the task not only broadly applicable but also substantially more challenging and motivates the need for new approaches capable of reasoning more effectively about visual-textual inconsistencies.

\subsection{Data}
\label{sec:bench_data}
For our MD-VQA protocol, we obtain training and testing examples by repurposing two recently proposed mistake datasets -  CaptainCook4D~\cite{peddi2024captaincook4d} and EgoPER~\cite{lee2024egoper}. CaptainCook4D contains 384 recordings (totaling 94.5 hours) that depict 24 high-level recipes, while EgoPER consists of 386 videos demonstrating the preparation of 5 recipes, totaling 28 hours of footage.
Both datasets include temporal annotations for fine-grained action steps corresponding to individual sub-steps within the high-level recipes.
Since our protocol focuses on localized fine-grained reasoning, we extract short video segments corresponding to individual action steps using the start and end timestamps provided in the dataset annotations. We next describe how we refactor each dataset for our task in question.

\textbf{CaptainCook4D}~\cite{peddi2024captaincook4d} comprises 350 types of fine-grained instructional steps \(t\), described in natural language. 
Each action \(t\) has a set of video clips \(\{v_k\}_{k=1}^{K_t}\), pre-extracted as described above, depicting either a correct or an incorrect execution with respect to the provided textual description.
If the step is performed incorrectly, the annotations specify a high-level mistake category \(e\); otherwise, the clip is labeled as correct.
For our problem, we retain 5 high-level mistake classes from CaptainCook4D’s original 7 categories: \textit{Preparation}, \textit{Technique}, \textit{Timing}, \textit{Measurement}, and \textit{Temperature}. 
We exclude the \textit{Ordering} and \textit{Missing} error types, as these pertain to multiple segments rather than single clips. Since the annotations in CaptainCook4D are noisy, we manually corrected the descriptions to make sure the corresponding clips are mistakes with respect to the instructions, and that they are grammatically correct. 
We provide the full details in the supplementary material. The resulting version of CaptainCook4D processed in this way is referred to as \textbf{CC-VQA}.

\textbf{EgoPER~\cite{lee2024egoper}} follows a different annotation format, where correct and erroneous steps are not considered as two variants of the same instructional step. Instead, they are treated as two separate fine-grained steps, $t \text{ and } t'$, without an established correct-mistake correspondence. 
For consistent evaluation across both datasets, we extend the original EgoPER annotations. Specifically, while the normal steps $t$ are left unmodified, for each mistake step $t'$ we select a corresponding normal step $t$ that specifies how $t'$ would appear if no mistake were made. For example, for a mistake step $t'$ (\textit{stir using knife}), we provide a corresponding normal step $t$ (\textit{stir using spoon}) and assign the mistake to a high-level mistake category $e$ (\textit{utensil error}).

Overall, we define four new mistake categories: \textit{Slip}, \textit{Technique}, \textit{Measurement}, and \textit{Utensil}.  
\textit{Slip} mistakes arise from carelessness (e.g., \textit{spilling water}).  
\textit{Utensil} mistakes occur when an incorrect tool is used (e.g., \textit{stirring with a knife} instead of a spoon).  
\textit{Measurement} mistakes involve incorrect quantity or timing.  
Finally, \textit{Technique} mistakes result from imprecise execution of an action (e.g., \textit{folding a tortilla into a quarter-circle} instead of a half-circle).
As with CaptainCook4D, we annotate video clips \(v_k\) which show an incorrect execution of step \(t\) with one of these mistake types depending on the description $t'$, while clips showing correct executions are labeled as correct. A complete mapping of mistake steps $t'$ to normal instruction steps \(t\) as well as the corresponding error types \(e\) is provided in the supplementary material. 
The resulting version of Ego-PER processed in this way is referred to as \textbf{EP-VQA}.

\begin{table}[tbp]
    \centering
    \resizebox{0.6\columnwidth}{!}{%
    \begin{tabular}{l|cc|cc|cc}
        \topline
        \textbf{Dataset} & \multicolumn{2}{c|}{\textbf{Train}} & \multicolumn{2}{c|}{\textbf{Test Seen}} & \multicolumn{2}{c}{\textbf{Test Unseen}} \\
        & \scriptsize{ Normal } & \scriptsize{ Error } & \scriptsize{ Normal } & \scriptsize{ Error }  & \scriptsize{ Normal } & \scriptsize{ Error }  \\
        \hline
        EP-VQA & 3571 & 270 & 602 & 54 & 532 & 60 \\  
        CC-VQA & 3186 & 1077 & 484 & 107 & 459 & 100 \\ 
        \bottomline
    \end{tabular}
     }
    \caption{\textbf{Number of samples for the EP-VQA and CC-VQA datasets.}}
    \label{tab:splits}
    %\vspace{-1em}
\end{table}

%\vspace{-1em}
\subsection{Evaluation}
\label{sec:bench_eval}
%\vspace{-1pt}
We divide the processed samples for both {CC-VQA} and {EP-VQA} into training and test sets. For both datasets, we define three subsets: train, test seen, and test unseen. 
The \textbf{unseen test set} (\(D_{\text{unseen}}\)) contains video clips from high-level activities (recipes) that were not included in the training set. This ensures that the instructional steps in this split have not been observed during training, allowing us to assess the generalization capability of the tested models to previously unseen textual descriptions.
The remaining portion of the dataset \(D' = D \setminus D_{\text{unseen}}\) is then further split into the \textbf{train set} (\(D_{\text{train}}\)) and the \textbf{seen test set} (\(D_{\text{test}}\)). The detailed statistics of the resulting splits are provided in Tab.~\ref{tab:splits}. 
We use both test sets to evaluate the performance of VLMs for video mistake detection in the MD-VQA setup
as described in Sec.~\ref{sec:bench_task}. Methods are evaluated by their recall, precision, and F1-score, computed with mistakes as the positive class.

\section{Method}
\label{sec:method}
%\vspace{-1pt}
For a model to effectively address the above-defined MD-VQA task, it needs to be able to perform fine-grained reasoning over the textual description \(t\), the video \(v\), and their interplay. For instance, a video showing a person preheating an oven to \(150^\circ\mathrm{C}\) should be considered incorrect for the instruction \textit{Preheat the oven to \(200^\circ\mathrm{C}\)}, despite the close semantic similarity of those two instructions.
To address this challenge, we draw inspiration from the success of reinforcement learning-based post-training across several tasks that require high-level reasoning. Building on these insights, we propose a novel post-training technique to enhance the video mistake recognition capabilities of VLMs. Specifically, we introduce a reward function designed to more accurately capture mistakes in the execution of instructional steps.
In the following sections, we provide a detailed overview of our proposed approach. 

\subsection{Preliminaries}
\label{sec:grpo}
GRPO~\cite{shao2024deepseekmath} is a reinforcement learning-based algorithm designed to enhance the reasoning capabilities of underlying models. Unlike earlier post-training algorithms~\cite{schulman2017proximal}, GRPO does not rely on a separately trained model to rank the policy network’s responses. Instead, it uses predefined rule-based rewards that are assigned to a group of candidate responses sampled from the policy model \( \pi_{\theta} \).

More precisely, the GRPO objective is computed by first sampling a group of \( G \) different responses \( \{o_i\}_{i=1}^G \) from the policy model \( \pi_{\theta} \). For each response \( o_i \), a reward value \( r_i \) is assigned using predefined reward functions. These rewards, which reflect the quality of the corresponding responses, are then used to compute per-response advantage values relative to the other candidates within the group. Formally:

\begin{equation}
A_i = \frac{r_i - \text{mean}(r_1, \dots, r_G)}{\text{std}(r_1, \dots, r_G)}.   
\label{eq:grpo_advantage}
\end{equation}

\noindent The policy model \( \pi_{\theta} \) is then optimized using the following objective:
\begin{equation}
\begin{aligned}
J_{\text{GRPO}}(\theta)  &= \frac{1}{G} \sum_{i=1}^{G}  
\min \!\Big( 
l_i A_i, \text{clip } \!\big( 
l_i, \; 1-\epsilon, \; 1+\epsilon 
\big) A_i 
 \Big) - \beta \,  \mathbb{D}_{\text{KL}}\!\big[ \pi_{\theta} \;||\; \pi_{\text{ref}} \big], \\
 l_i &= \frac{\pi_{\theta}(o_i \mid q, v)}{\pi_{\theta_{\text{old}}}(o_i \mid q, v)},
\end{aligned}
\label{eq:grpo_obj}
\end{equation}
where $q$ is a question prompt concerning a video snippet $v$ and  $\pi_{\theta_{\text{old}}}$ is the model from the previous iteration.
In this objective, the first term encourages the policy video-language model \( \pi_\theta \) to generate responses with higher advantages, while the second Kullback-Leibler divergence term acts as a regularizer that prevents excessive deviations of the optimized policy model \( \pi_\theta \) from the reference policy \( \pi_{\text{ref}} \). The reference policy model \( \pi_{\text{ref}} \) is initialized from the instruction-tuned model and kept fixed during training.

\begin{figure*}[t!]
    \centering
    \includegraphics[width=\linewidth]{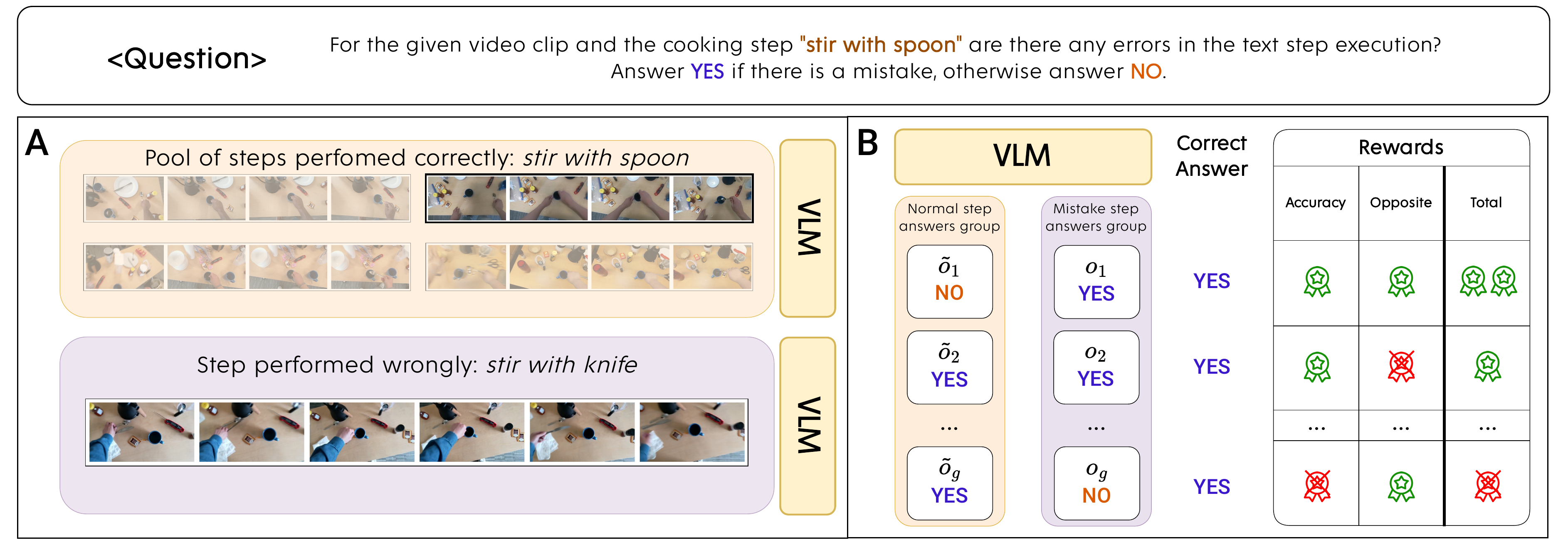}
    \caption{\textbf{Overview of our method.} 
    To improve the reasoning capabilities of VLMs for mistake detection, we post-train models with GRPO using a novel \textit{opposite} reward function, which encourages different responses for video clip pairs demonstrating correct and incorrect executions of the same step. A: We pair an incorrectly performed step (bottom) with an ``opposite'' correct execution from a reference pool (top). The VLM processes both videos independently. B: The model earns an accuracy reward for correct answers, and an opposite reward if its answers for the two clips differ from one another. The opposite reward is only given if the accuracy reward is achieved. Through this training, the model's ability to detect mistakes as subtle discrepancies between instructional step descriptions and their execution is improved.
    }
    \label{fig:method}
    \vspace{-0.3cm}
\end{figure*}

\subsection{Reward Functions for MD-VQA}
\label{sec:rew}
To address the task of mistake detection effectively, we utilize a combination of three reward functions for relative advantage estimation: \textit{format} and \textit{accuracy} rewards as in~\cite{guo2025deepseek}, as well as our novel \textit{opposite} reward function. 

\paragraph{\textbf{Format reward}.} It measures whether a response adheres to a specific predefined output format. Formally, for a response \(o_i\) and a predefined \(\text{format}\) string, this reward is computed as:
\[
r_i^{\text{format}} = \mathbf{1}[\text{format}(o_i)],
\]
where $\mathbf{1}[\cdot]$ is the indicator function.
In our work, we adopt the ``thinking'' format, in which reasoning traces are enclosed between the \texttt{<think></think>} tags, and the final answer is provided between the \texttt{<answer></answer>} tags. Enforcing this format encourages the model to perform intermediate reasoning steps, allowing it to process textual and visual tokens in greater detail before producing the final answer to a given question.

%\vspace{-1pt}
\paragraph{\textbf{Accuracy reward}.} It measures the correctness of the responses with respect to the provided ground-truth answers. Specifically, for a response \( o_i \), we first extract the predicted answer $\corrans_i$ enclosed between the \texttt{<answer></answer>} tags. The accuracy reward with respect to the target answer $\gtans$ is then defined as:  
\[
r_i^{\text{accuracy}} = \mathbf{1}[\corrans_i = \gtans],
\]
where equality holds if the texts $\corrans_i$ and $\gtans$ match exactly. This reward encourages the model to produce factually correct responses, which corresponds to correctly determining whether a given text--video pair contains a mismatch or is mistake-free.

%\vspace{-1pt}
\paragraph{\textbf{Opposite reward}.}
\label{sec:opp_reward}
In addition to the above functions, we propose a novel reward specifically designed to enhance the model's ability to identify discrepancies between textual and visual information, as illustrated in Fig.~\ref{fig:method}. 
Depending on the segment \(v\), it may depict a correct or incorrect execution of the step $t$. We then can also sample an ``opposite'' video segment \(\tilde{v}\) which represents the \textit{opposite} scenario; that is, if \(v\) shows a correct execution of step \(t\), \(\tilde{v}\) is sampled to show a failed execution, and vice versa. 
Intuitively, a policy model sensitive to mismatches between textual and visual information should produce \textit{dissimilar} responses for the same question \(q\) about step \(t\) when given these two video variants. Formally, if $\corrans_i$ and $\oppans_i$ are the extracted answers from responses $o_i$ and $\tilde{o}_i$ for question \(q\) based on video segments \(v\) and \(\tilde{v}\), respectively, the opposite reward is
\[
\mathbf{1}[\corrans_i \neq \oppans_i].
\]
To prevent this reward from incentivizing factually incorrect responses, \eg, classifying a correct segment as incorrect and vice versa, we additionally constrain it using a multiplicative accuracy gate. As such, the final opposite reward is:
\[
r_i^{\text{opposite}} = \mathbf{1}[\corrans_i \neq \oppans_i] \cdot \mathbf{1}[\corrans_i = \gtans],
\]
where $\gtans$ is the ground-truth answer to the question $q$. This reward, together with the previous two functions, encourages the policy model to be sensitive to visual changes that correspond to violations of the textual descriptions.

\paragraph{\textbf{Final reward}.}
As our final reward function, we utilize a combination of the above-described rewards:
\[
    r_i = r_i^{\text{accuracy}} + r_i^{\text{format}} + \lambda {r}_i^{\text{opposite}},
\]
where $\lambda$ balances the influence of our proposed \textit{opposite} reward function on the final outcome. We evaluate the impact of $\lambda$ in our experiments.

\subsection{Training}
\label{sec:grpo_train}
For the \textit{opposite reward} computation, each mistake segment requires a corresponding normal example, and vice versa. 
Given a text–video pair $(v, t)$ containing a mistake, we randomly sample a normal video clip $\tilde{v}$ with the same step description $t$. Similarly, for a normal video clip, we sample a corresponding mistake video segment that depicts the same instructional step. In the rare cases where no video with a wrong execution for text $t$ exists, we randomly select a mistake video clip $\tilde{v}'$ from any unrelated step $t'$. We emphasize that this sampling step is \emph{not} required during the evaluation phase.
We directly apply GRPO with our reward functions to the VLMs and do not rely on an initial fine-tuning stage, similar to~\cite{liu2026videoauto}.
%-------------------------------------------------------------------------

\section{Experiments}
\label{sec:exp}

We evaluated a diverse set of baselines on our benchmark to thoroughly compare our proposed approach with existing methods:

\begin{itemize}
    \item \textbf{Zero-Shot} (ZS): base VLM model without additional training as in~\cite{peddi2024captaincook4d}.
    
    \item \textbf{Zero-Shot + Reasoning} (ZS + \textit{R}): the base VLM model, where we extend the prompt with an explicit request for structured reasoning. The model is instructed to provide its reasoning within \texttt{<think>...</think>} tags and its final answer within \texttt{<answer>...</answer>}. 
    
    \item \textbf{Supervised Fine-Tuning} (SFT): the base VLM model fine-tuned with the standard CE loss using the training set of question–answer pairs described in Sec.~\ref{sec:bench_task}. 
    
    \item \textbf{Supervised Fine-Tuning + Reasoning}  (SFT + \textit{R}): the SFT model fine-tuned as described above, but during inference queried with a reasoning prompt, as in the (ZS + R) baseline. 

    \item \textbf{Supervised Fine-Tuning with Explanation} (SFTe): the SFTe model is trained not only for mistake detection but also for error explanation, similar to~\cite{patsch2025ICCV}: in addition to predicting if there is a mistake or not, the model has to explain what the error is. For example, \textit{YES, step `stir using spoon' is performed incorrectly. The video shows step `stir using knife'.} 
    
    \item \textbf{GRPO}: the base VLM model post-trained using the original GRPO objective with accuracy and format rewards, without incorporating our proposed opposite reward.
\end{itemize}
    
Additionally, we report experiments on the EP-VQA unseen split with the commercial GPT-4o-mini model~\cite{achiam2023gpt} using the reasoning prompt.
For the baselines evaluated with the reasoning prompt, as well as GRPO and our proposed method, we utilize the same prompt format, which we provide in the supplementary material. To ensure fairness, all evaluated methods share the identical base architecture.

%\vspace{-1pt}
\paragraph{Implementation details.}
Our base VLM model is Qwen2.5-VL-7B~\cite{bai2025qwen2}, fine-tuned via our proposed post-training approach described in Sec.~\ref{sec:method}. 
We follow~\cite{feng2025videor1} for our frame sampling strategy and resolution, as well as for the GRPO objective computation, setting the response group size $G$ to 8 and the KL-divergence weight to 0.04. 
For the opposite reward function, we use a balancing weight of $\lambda = 0.2$. Training is conducted with a batch size of 4 on 4 NVIDIA H100 GPUs. During training, we balance the number of mistakes and correct examples by resampling videos corresponding to mistake steps, since these are under-represented in the training sets of both EP-VQA and CC-VQA, with roughly 1:13 and 1:3 ratios of mistakes to normal examples, respectively. This resampling is applied for all trainable baselines. We found that without such resampling, the models become overly biased toward predicting the absence of mistakes due to the dataset imbalance, particularly for EP-VQA. Training dynamics and further details for all methods are provided in the supplementary material.

\begin{table*}[t]
    \centering
    \renewcommand{\arraystretch}{1.15}
    \setlength{\tabcolsep}{5pt}
    \resizebox{\textwidth}{!}{
    \begin{tabular}{l|ccc|ccc|ccc|ccc}
        \topline
        & \multicolumn{6}{c|}{\textbf{CC-VQA}} & \multicolumn{6}{c}{\textbf{EP-VQA}} \\
        \cline{2-13}
        \textbf{Method} & \multicolumn{3}{c|}{\textbf{Seen}} & \multicolumn{3}{c|}{\textbf{Unseen}} & \multicolumn{3}{c|}{\textbf{Seen}} & \multicolumn{3}{c}{\textbf{Unseen}} \\
        \cline{2-13}
        & \textit{Rec} & \textit{Prec} & \textbf{\textit{F1}} & \textit{Rec} & \textit{Prec} & \textbf{\textit{F1}} & \textit{Rec} & \textit{Prec} & \textbf{\textit{F1}} & \textit{Rec} & \textit{Prec} & \textbf{\textit{F1}} \\
        \hline
        ZS & 0.0 & 0.0 & 0.0 & 0.0 & 0.0 & 0.0 & 0.0 & 0.0 & 0.0 & 0.0 & 0.0 & 0.0 \\
        ZS + \textit{R} & 41.9 & 16.7 & 23.8 & 54.8 & 20.6 & 29.9 & 60.4 & 12.5 & 20.7 & 21.7 & 12.3 & 15.7 \\
        GPT-4o-mini + \textit{R} & - & - & - & - & - & - & - & - & - & 50.0 & 20.4 & 29.0 \\
        SFT & 43.0 & \textbf{21.5} & 28.7 & 60.0 & \textbf{24.2} & \underline{34.5} & 1.9 & 14.3 & 3.3 & 3.3 & 7.7 & 4.7 \\
        SFT + \textit{R} & 54.2 & 17.7 & 26.7 & 64.0 & 19.3 & 29.7 & 24.1 & 7.6 & 11.6 & 6.7 & 10.5 & 8.2 \\
        SFTe & 49.5 & 19.6 & 28.2 & 55.0 & 20.6 & 30.0 & 3.7 & \underline{40.0} & 6.8 & 0.0 & 0.0 & 0.0 \\
        GRPO & \textbf{66.4} & \underline{20.2} & \textbf{30.9} & \underline{71.0} & 21.2 & 32.6 & \underline{74.1} & 39.2 & \underline{51.3} & \underline{66.7} & \underline{31.7} & \underline{43.0} \\
        \hline
        \rowcolor{orange!30!white} \textbf{Ours} & \underline{61.7} & 19.6 & \underline{29.7} & \textbf{80.0} & \underline{22.8} & \textbf{35.5} & \textbf{79.6} & \textbf{40.6} & \textbf{53.8} & \textbf{70.0} & \textbf{36.5} & \textbf{48.0} \\
        \bottomline
    \end{tabular}
    }
    \caption{\textbf{Comparative evaluation on the CC-VQA and EP-VQA seen and unseen test splits.} Our proposed post-training method learns a generalizable mistake representation that performs well for both seen and unseen videos.}
    \label{tab:main_res}
    % %\vspace{-0.3cm}
    %\vspace{-1pt}
\end{table*}

%\vspace{-1pt}
\subsection{Results}
We provide the main results on the seen and unseen splits of {CC-VQA} and {EP-VQA} in Tab.~\ref{tab:main_res}.
To ensure a fair comparison, we verified that the different models' answers are incorrect due to wrong predictions rather than formatting issues. 
We find that the zero-shot model already reliably follows formatting instructions. For fine-tuned models, correct formatting is learned as part of the mistake detection task, and all outputs during inference conform to the required answer format. As a result, any differences in results indeed stem from differences in mistake detection performance.

\textbf{On {CC-VQA}}, the zero-shot approach~\cite{peddi2024captaincook4d} fails to detect any discrepancies and predicts all video--text pairs as matching, \ie, as correct executions. Adding a reasoning prompt (ZS + R) substantially boosts performance.
In contrast, SFT reaches strong performance for the seen and unseen splits without a reasoning prompt.
Requesting an additional explanation (SFTe) as proposed in~\cite{patsch2025ICCV} does not improve over standard SFT either. 
When it comes to RL-based methods, GRPO performs better than SFT on the seen split but not on the unseen split. However, our method consistently outperforms SFT for both splits. In particular, the recall is much higher, \ie, more errors are correctly detected, whereas the number of false positives is only slightly increased, as indicated by the slightly lower precision. Compared to GRPO, our approach reaches slightly lower results on the seen split, but it achieves $2.9$ points higher F1-score for the unseen split. This gain highlights the effectiveness of our \textit{opposite} reward design in learning a general notion of mistakes, which transfers better to unseen instructions. 

We additionally provide the per-error recall on the different mistake categories in Tab.~\ref{tab:c4d_per_mistake}.
For the seen steps, our method achieves the highest recall for two of the categories, and second-best on the remaining three, showcasing its consistency. 
For the unseen steps, our method achieves the best results across all mistake types with the exception of the \textit{temperature error}, which has only very few samples, confirming its success in transferring its mistake representation to unseen steps. The largest recall gain is observed for \textit{technique errors}, where it is $21.4$ points better than the second-best performing models GRPO and SFTe. 

\begin{table*}[t!]
    \renewcommand{\arraystretch}{1.05}
    \setlength{\tabcolsep}{4pt}
    \centering
    \resizebox{\textwidth}{!}{%
    \begin{tabular}{l|cc|cc|cc|cc|cc}
          \topline
            \multicolumn{1}{c|}{\multirow{2}{*}{\rule{0pt}{2.0ex}\textbf{Method}}} &
            \multicolumn{2}{c|}{\rule{0pt}{2.6ex}\textit{Technique (424)}} &
            \multicolumn{2}{c|}{\rule{0pt}{2.6ex}\textit{Preparation (339)}} &
            \multicolumn{2}{c|}{\rule{0pt}{2.6ex}\textit{Temperature (48)}} &
            \multicolumn{2}{c|}{\rule{0pt}{2.6ex}\textit{Measurement (276)}} &
            \multicolumn{2}{c}{\rule{0pt}{2.6ex}\textit{Timing (138)}} \\
            \cline{2-11}
            & \multicolumn{1}{c}{\scriptsize{\textbf{S} (46)}} & \multicolumn{1}{c|}{\scriptsize{\textbf{U} (28)}} &
              \multicolumn{1}{c}{\scriptsize{\textbf{S} (29)}} & \multicolumn{1}{c|}{\scriptsize{\textbf{U} (26)}} &
              \multicolumn{1}{c}{\scriptsize{\textbf{S} (11)}} & \multicolumn{1}{c|}{\scriptsize{\textbf{U} (6)}} &
              \multicolumn{1}{c}{\scriptsize{\textbf{S} (23) }} & \multicolumn{1}{c|}{\scriptsize{\textbf{U} (32)}} & 
              \multicolumn{1}{c}{\scriptsize{\textbf{S} (19)}} & \multicolumn{1}{c}{\scriptsize{\textbf{U} (20)}} \\
            \hline
            ZS &   0.0 & 0.0 & 0.0 & 0.0 & 0.0 & 0.0 & 0.0 &  0.0 & 0.0 & 0.0 \\
            ZS + \textit{R  } & \underline{39.1} & 26.1 & 51.7 & 56.0 & 40.0 & 33.3 & 54.5 & 63.3 & 50.0 & 65.0 \\
            SFT & 30.4 & 46.4 & 48.3 & \underline{69.2} & \underline{72.7} & \underline{66.7} & 69.6 & 65.6 & 52.6 & \textbf{75.0} \\
            SFT + \textit{R  } &  45.7 & 28.6 & 51.7 & \underline{69.2} & \underline{72.7} & \textbf{100.0} &  60.9 & 75.0 & 68.4 & 80.0 \\
            SFTe & 	 34.8 & \underline{50.0} & 55.2 & 61.5 & 63.6 & 0.0 & \textbf{82.6} & 75.0 & 57.9 & 30.0 \\
            GRPO & \textbf{56.5} & \underline{50.0} &  \textbf{75.9} & \underline{69.2} & 63.6 & \underline{66.7} & \underline{78.3} & 84.4 & \textbf{78.9} & \underline{70.0} \\
            \hline
            \rowcolor{orange!30!white} \textbf{Ours} &  \underline{39.1} & \textbf{71.4} & \underline{69.0} & \textbf{76.9} & \textbf{90.9} & 33.3 & \textbf{82.6} & \textbf{96.9} & \underline{73.7} & \textbf{75.0} \\
          \Xcline{1-11}{0.8pt}
    \end{tabular}
    }
    \vspace{6pt}
    \caption{\textbf{Recall for the seen and unseen test splits of CC-VQA.} We show the number of samples for each mistake type in parentheses for seen (S) and unseen (U) splits. The number of training error samples is shown next to the name of the corresponding mistake category. Our method consistently scores among the best methods in mistake recall for both seen and unseen instructions.}
    \label{tab:c4d_per_mistake}
    %\vspace{-0.2cm}
\end{table*}

\textbf{On EP-VQA}, the zero-shot methods exhibit similar trends. However, for the other methods, some clear differences emerge.
The SFT models underperform on this benchmark and struggle to recognize errors, predicting mostly that there are no mistakes present in the clips. Since EP-VQA contains way fewer mistake instances than CC-VQA, we hypothesize that these models quickly overfit to the mistake examples, despite the mistake upsampling applied, and therefore fail to generalize to the test splits.
Post-training methods, on the other hand, perform very well: GRPO achieves $51.3$ and $43.0$ F1-score on the seen and unseen splits, respectively, with our method in turn improving upon GRPO by $2.5$ and $5$ points. Again, the gap is larger on the unseen split, showing that our learned mistake representation generalizes better to new steps and recipes.
We provide the per-error analysis on the EP-VQA dataset, showing similar trends to the CV-VQA, in the supplementary material, where we also show results with an additional model. 

To place these results in a broader context and investigate how they compare to closed-source models, we tested the commercial GPT-4o-mini model on the unseen split of EP-VQA as well. We find that it outperforms the base Qwen2.5 model by $13.3$ points in F1 score when using the same reasoning prompt. However, when we adapt the base model with our post-training approach, the performance jumps to $48.0$, which is by $19.0$ points higher than the commercial model, showing the benefits of mistake-specific post-training.

\begin{figure*}[tb]
    \centering
    \includegraphics[width=\linewidth]{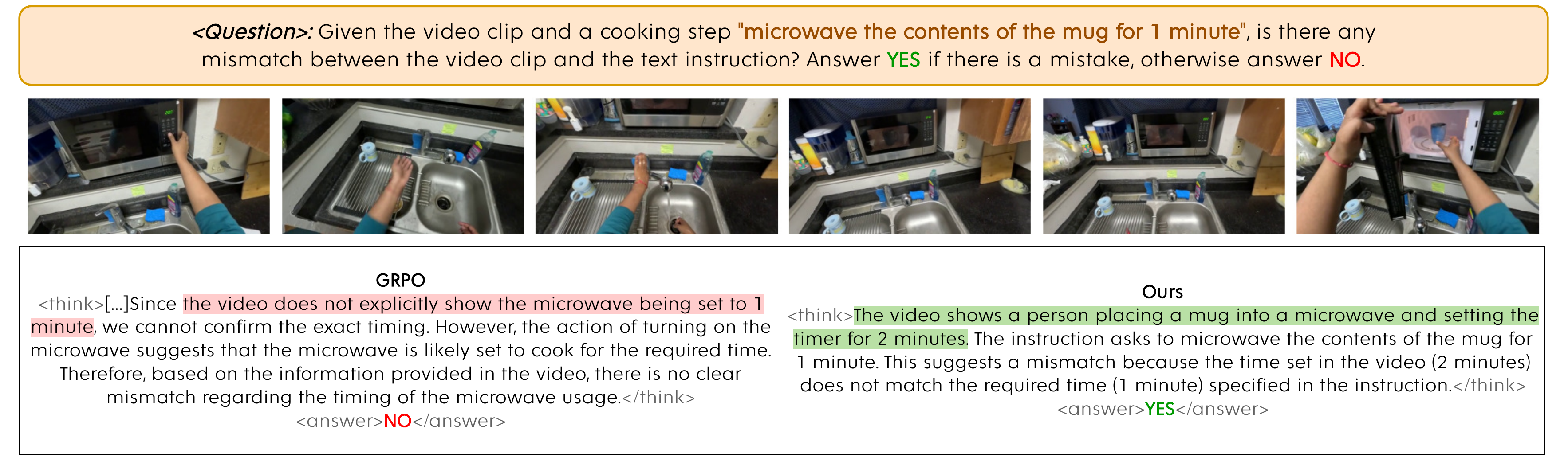}
    \caption{\textbf{Qualitative example on the CC-VQA Unseen split.} We show a clip for the \textit{spiced hot chocolate} activity. The instruction specifies microwaving the milk for one minute, but the video shows the user setting the timer to two minutes. The standard GRPO baseline (left) misses this fine-grained visual cue and assumes the action is correct, whereas our method (right) grounds its reasoning in the visual evidence (detecting the press of the “2” button in the first frame) and correctly identifies the mistake.}
    %   
    %\vspace{-0.4cm}
    \label{fig:qual}
\end{figure*}

\begin{figure*}[tb]
    \centering
    \includegraphics[width=\linewidth]{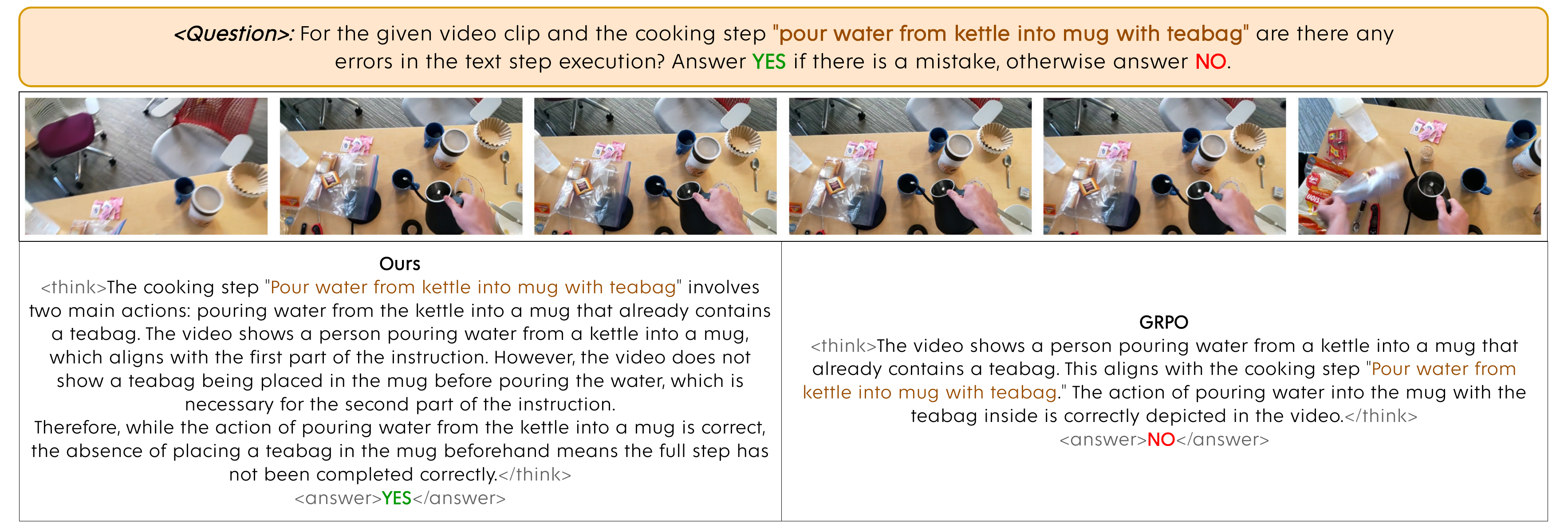}
    \caption{\textbf{Qualitative example on the EP-VQA Seen split.} The video shows the step \textit{pour water from a kettle into a mug}. The top mug contains a teabag, while the bottom one does not. The mistake arises from pouring water into the mug without the teabag. Our model (left) successfully identifies this subtle discrepancy, outperforming the standard GRPO (right).}
    \label{fig:egoper-seen-qual}
    \vspace{-0.5cm}
\end{figure*}

In general, we find that VLMs come with two main limitations for the MD-VQA task. First, there is a bias towards ``no error". We believe that mistake-specific training strategies, such as our opposite reward, can help remedy this, as evidenced by the improved performance.
Second, there is the issue of hallucinations. For example, in some cases the reasoning traces do not mention errors, yet the final answer is ``mistake". 
While this is a limitation of LLM-based models as a whole, this could be mitigated with a consistency loss or reward.  
Overall, despite the consistent gains demonstrated by our approach over the baselines, the absolute performance on the proposed benchmark leaves plenty of room for improvement, underscoring the inherent complexity of the task and highlighting that existing methods still require considerable improvement to address its challenges. 

\subsection{Qualitative examples}

Fig.~\ref{fig:qual} and Fig.~\ref{fig:egoper-seen-qual} show two qualitative examples of our proposed approach. 
Fig.~\ref{fig:qual} depicts an example from the unseen CC-VQA test dataset with a fine-grained duration mismatch: the microwave timer is set to two minutes instead of the instructed one minute. By being trained to detect such precise visual discrepancies, our method successfully spots the "2" button being pressed, recognizes that the video does not match the step description, and accurately labels it as a mistake. In contrast, standard GRPO overlooks this subtle cue, incorrectly assuming the timing aligns with the text and failing to detect the mistake. 
In Fig.~\ref{fig:egoper-seen-qual}, 
the participant is instructed to pour water from the kettle into the mug containing the teabag (top-right mug), but instead pours it into the central mug without a teabag. Both GRPO and our method correctly detect that the first part of the step (“pour water from kettle”) is executed properly, but our approach additionally identifies that the second part (“into mug with teabag”) is violated.
More qualitative examples can be found in the supplementary material.

\subsection{Ablations}
\textbf{Effect of $\lambda$.}
We test the effect of varying $\lambda$, the weight of our \textit{opposite} reward, in Tab.~\ref{tab:lambda_macroF1_seen_unseen} for both seen and unseen test splits of the {EP-VQA} dataset. We find that a value of $\lambda=0.2$ yields the best performance on both splits. In general, using the opposite reward is always beneficial for generalizability to unseen actions.  We fix $\lambda=0.2$ throughout our experiments. %, and improves results for values below $0.3$.

\begin{table}[tbp]
    \centering
    % \resizebox{0.8\linewidth}{!}{%
    \begin{tabular}{c|c|c}
        \topline
        \rule{0pt}{2.4ex} \hspace{0.15cm} $\lambda$ \hspace{0.15cm} & { Seen (F1) } & { Unseen (F1) } \\ [0.2em]
        \hline
        0.0 & 51.3 & 43.0 \\
        0.1 & 53.0 & 45.7 \\
        \rowcolor{orange!30!white} 0.2 & \textbf{53.8} & \textbf{48.0} \\
        0.3 & 44.9 & 43.3 \\
        \bottomline
    \end{tabular}
    % }
    % \vspace{6pt}
    \caption{\textbf{Effect of $\lambda$.} We show the F1 for seen and unseen test splits on EP-VQA.}
    \label{tab:lambda_macroF1_seen_unseen}
\end{table}

% \vspace{-1em}
\paragraph{\textbf{Opposite reward formulation.}}
Our opposite reward formulation includes the accuracy gate ($\mathbf{1}[\corrans_i = \gtans]$) that ensures it only provides a reward if the original question is correctly answered. We ablate this choice in Tab.~\ref{tab:ropp_macroF1_seen_unseen}, where we find that the accuracy gate improves performance by $3.5$ F1 on the seen split and $4.4$ F1 on the unseen split. Conceptually, if the predicted answer for the original question is wrong, enforcing this opposite reward will only confuse the model, which is why this term is beneficial.

\begin{table}[tbp]
    \centering
    \begin{tabular}{c|c|c}
        \topline
        \rule{0pt}{2.4ex} Method & { Seen (F1) } & { Unseen (F1) } \\ [0.2em]
        \hline
        \rowcolor{orange!30!white} w. acc.   & \textbf{53.8} & \textbf{48.0}   \\
        w/o acc. & 50.3 & 43.6  \\
        \bottomline
    \end{tabular}
    % \vspace{6pt}
    \caption{\textbf{Ablating $r_i^{opp}$ formulation.} We show the F1 for both test splits on EP-VQA. }
    \label{tab:ropp_macroF1_seen_unseen}
\end{table}

\paragraph{\textbf{Post-training parameters.}} We analyze the effect of the GRPO parameters $G$ (group size) and $\beta$ (KL-divergence weight) on the performance of our method. We show the results of these ablations in Tab.~\ref{tab:group_size} and Tab.~\ref{tab:kl_weight}, respectively.
We find that reducing the group size $G$, which controls how many responses the model generates for advantage estimation, consistently degrades performance, with the F1 score on both splits dropping by more than 20 points for $G=2$. While increasing the group size beyond $G=8$ could potentially improve performance further, this comes at a higher computational cost, so we limit $G$ to 8 to reduce the overall computational burden of our method.

For the KL-term weight $\beta$, which prevents the policy model from deviating too far from the instruction-tuned model $\pi_{ref}$, we use a value of $\beta=0.04$. Reducing the value to $\beta=0.02$, which allows greater deviation from the reference policy, slightly worsens performance, whereas increasing the constraint on deviation with $\beta=0.06$ leads to a more substantial drop in performance. Therefore, we follow common protocol (e.g.,~\cite{feng2025videor1}) and fix $\beta=0.04$.

\begin{table}[h!]
\centering
\begin{minipage}{0.45\linewidth}
\centering
\vspace{6pt}
\resizebox{\linewidth}{!}{%
    \begin{tabular}{c|c|c}
    \topline
    \rule{0pt}{2.4ex} \hspace{0.15cm} $G$ \hspace{0.15cm} & { Seen (F1) } & { Unseen (F1) } \\ [0.2em]
    \hline
    2 & 32.7 & 22.2  \\
    4 & 37.4 & 39.2  \\
    6 & 42.1 & 41.6 \\
    \rowcolor{orange!30!white} 8 & \textbf{53.8} & \textbf{48.0} \\
    \bottomline
    \end{tabular}
}
\caption{\textbf{Effect of group size $G$.} We show the F1 for seen and unseen test splits on EP-VQA.}
\label{tab:group_size}
\end{minipage}
\hfill
\begin{minipage}{0.45\linewidth}
\centering
\vspace{6pt}
\resizebox{\linewidth}{!}{%
    \begin{tabular}{c|c|c}
    \topline
    \rule{0pt}{2.4ex} \hspace{0.15cm} $\beta$ \hspace{0.15cm} & { Seen (F1) } & { Unseen (F1) } \\ [0.2em]
    \hline
    0.02 & 51.2 & 42.0   \\
    \rowcolor{orange!30!white} 0.04 & \textbf{53.8} & \textbf{48.0}  \\
    0.06 & 45.0 & 41.4 \\
    \bottomline
    \end{tabular}
}
\caption{\textbf{Effect of the KL-term weight.} We show the F1 for seen and unseen test splits on EP-VQA.}
\label{tab:kl_weight}
\end{minipage}
\end{table}

% \vspace{-1em}
\paragraph{\textbf{Limitations}}
The clip-level descriptions in CC-VQA and EP-VQA do not always capture the broader procedural context. Consequently, mistakes that depend on earlier or later steps may be ambiguous or impossible to identify from an isolated video--text pair. Incorporating longer temporal context would enable the evaluation of such dependencies. Moreover, although our unseen splits contain held-out activities, they originate from the same datasets and domains as the training data. Generalization to substantially different procedures, recording conditions, and mistake types remains to be established.

%-------------------------------------------------------------------------
\section{Conclusion}

In this work, we introduced the MD-VQA protocol and benchmark for stepwise mistake detection in videos. MD-VQA is specifically designed to assess the generalizability of methods to both seen and unseen steps, reflecting real-world scenarios.
To address this task, we proposed the first VLM post-training method for video mistake detection. By designing a custom reward function, our method enables VLMs to successfully attend to subtle execution details that current models overlook.
We demonstrated that our approach achieves state-of-the-art performance against zero-shot, fine-tuning, and post-training baselines on MD-VQA, especially for steps never observed during training. 
% For instaince,, such as those in the unseen quesadilla recipe, we improve the F1 score by $11.6\%$ over the previous best results. 
These findings highlight the potential of our approach, and we hope our protocol will encourage future research in this important field.

%-------------------------------------------------------------------------
\section*{Acknowledgments}
The work has been supported by the ERC Consolidator Grant FORHUE (101044724) and the Federal Ministry of Research, Technology and Space (BMFTR) under grant no.\ 01IS22094A WEST-AI. For the computations involved in this research, we acknowledge EuroHPC Joint Undertaking for awarding us access to Leonardo at CINECA, Italy, through EuroHPC Regular Access Call - proposal No.\ EHPC-REG-2025R01-218.

\bibliography{egbib}

\appendix
% \documentclass{bmvc2k}

% %% Enter your paper number here for the review copy
% % \bmvcreviewcopy{211}

% \title{Post-Training VLMs for Video Mistake Detection\\Supplementary Material}

% % Enter the paper's authors in order
% \addauthor{Federico Spurio*}{fspurio@iai.uni-bonn.de}{1,2}
% \addauthor{Olga Zatsarynna*}{zatsarynna@iai.uni-bonn.de}{1,2}
% \addauthor{Lars Doorenbos*}{doorenbos@iai.uni-bonn.de}{1,2}
% \addauthor{Emad Bahrami}{bahrami@iai.uni-bonn.de}{1}
% \addauthor{Gianpiero Francesca}{gianpiero.francesca@gmail.com}{3}
% \addauthor{Juergen Gall}{gall@iai.uni-bonn.de}{1,2}

% % Enter the institutions
% % \addinstitution{Name\\Address}
% \addinstitution{
%  University of Bonn\\
%  Bonn, Germany
% }
% \addinstitution{
%  Lamarr Institute for Machine Learning and Artificial Intelligence\\
%  Bonn, Germany
% }
% \addinstitution{
%  Toyota Motor Europe\\
%  Belgium
% }

% \runninghead{Spurio et al.}{Post-Training VLMs for Video Mistake Detection}

% % Any macro definitions you would like to include
% % These are not defined in the style file, because they don't begin
% % with \bmva, so they might conflict with the user's own macros.
% % The \bmvaOneDot macro adds a full stop unless there is one in the
% % text already.
% \def\eg{\emph{e.g}\bmvaOneDot}
% \def\Eg{\emph{E.g}\bmvaOneDot}
% \def\etal{\emph{et al}\bmvaOneDot}

% %-------------------------------------------------------------------------
% % Document starts here
% \begin{document}

% \maketitle

\section{Implementation details}
\label{sec:supp_impl}
To ensure a fair comparison with the baselines, we adopt the same training hyperparameters across all methods whenever possible. We train for $800$ and $802$ iterations on EP-VQA and CC-VQA, respectively. Training is conducted with a batch size of $4$ on $4$ NVIDIA H100 GPUs. For the frame resolution and sampling strategy, we follow~\cite{feng2025videor1}.
Moreover, we perform upsampling of mistake instances during training, with upsampling factors of $13$ and $3$ for the EP-VQA and CC-VQA datasets, respectively. Without this balancing, the approaches become highly biased toward classifying video--text pairs as matching, with models trained on EP-VQA collapsing to always predict normal examples.

\paragraph{\textbf{GRPO}.}
We use the same hyperparameters as in our proposed post-training approach, but do not apply the proposed opposite reward. Instead, the model is trained using only the accuracy and format rewards. All other hyperparameters are identical to those used in our method. We utilize the reasoning prompt during both training and evaluation. 

\paragraph{\textbf{SFT and SFTe}.}
Since SFT and SFTe are trained using the standard CE loss (with YES/NO answers and YES/NO + explanation), GRPO-specific parameters do not apply, while the remaining hyperparameters are kept fixed. During training, we use the base prompt for these methods, whereas testing is performed either with the base prompt (SFT and SFTe) or with the reasoning prompt (SFT + \textit{R}).

\section{Additional Results}
\label{sec:supp_add_res}

\begin{figure}[h!]
    \centering
    \includegraphics[width=0.8\linewidth]{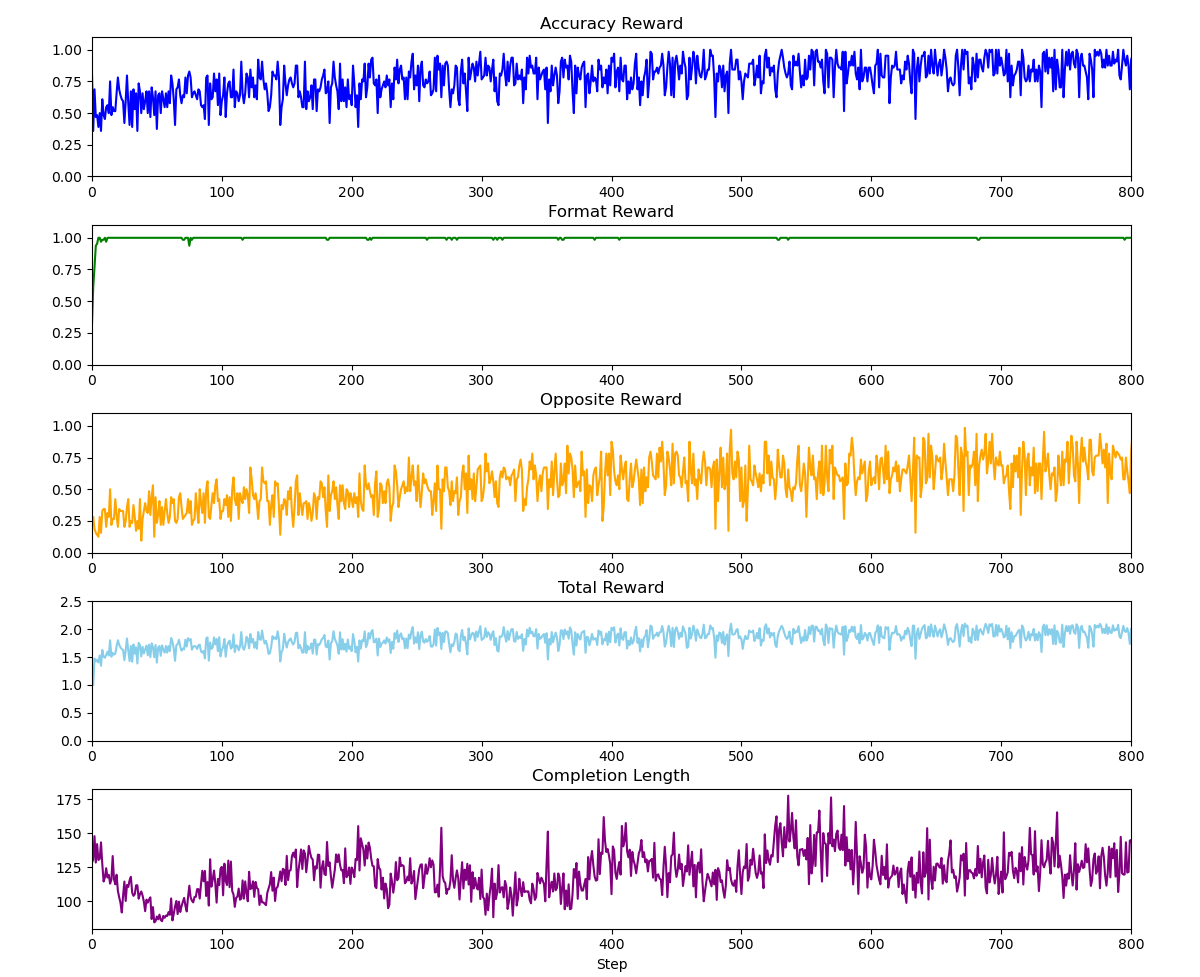}
    \caption{\textbf{Training metrics of our proposed post-training approach on EP-VQA.}}
    \label{fig:grpo_opp_train}
\end{figure}

% --------------------------- METRICS -----------------------------------
\paragraph{\textbf{Post-training metrics} on EP-VQA.}
In Fig.~\ref{fig:grpo_opp_train}, we demonstrate how the individual reward terms and the completion length of our proposed approach evolve during training. As can be seen, all reward components used in our method, namely the format, accuracy, and opposite rewards, consistently increase throughout the training process, indicating that the policy model learns to produce answers that satisfy the given reward criteria.

The format reward reaches its highest value very early in training, showing the inherent ability of Qwen2.5-VL to follow the required output format. In contrast, the accuracy and opposite rewards improve gradually.
We observe that the opposite reward remains lower than the accuracy reward for most training steps, showing that computing the opposite reward presents a more challenging task than the standard accuracy reward,
providing an additional informative training signal during fine-tuning. For the total reward computation, the opposite reward is incorporated with a weighting factor of $\lambda = 0.2$, as described in the main paper.

We also show the change in completion length throughout training. While there are some fluctuations at the beginning, toward the end of the training stage the completion length converges to an average of 125 tokens, which approximately corresponds to 90 to 100 words.

% --------------------------- PER-ERROR RESULTS  -----------------------------------
\paragraph{\textbf{Per-Error results} on EP-VQA.}
Similar to CC-VQA, we report per-error recall for the EP-VQA dataset in Tab.~\ref{tab:egoper_per_mistake}. The unseen split does not contain errors for the “Measurement” class, leading to empty entries in the corresponding column. 
Also here, our post-training method achieves the highest recall for nearly all cases. 
On the seen split, our method performs better than or on par with other baselines across all different mistake types. On the unseen split, our approach achieves a slightly lower recall than GRPO on the “Slip” error, but shows significantly higher recall for the “Utensil Error”, with an improvement of +12.5 percentage point. 
We observe, however, that all methods struggle with unseen “Technique” error instances. This highlights the difficulty of learning a general notion of mistakes that can be effectively generalized to steps unseen during training.

\begin{table*}[h!]
\centering
\renewcommand{\arraystretch}{1.05}
\setlength{\tabcolsep}{4pt}
\resizebox{0.8\textwidth}{!}{%
\begin{tabular}{l |cc|cc|cc|cc}
  \topline
    \multicolumn{1}{c|}{\multirow{2}{*}{\rule{0pt}{2.0ex}\textbf{Method}}} &
    \multicolumn{2}{c|}{\rule{0pt}{2.6ex}\textit{ Technique (55)}} &
    \multicolumn{2}{c|}{\rule{0pt}{2.6ex}\textit{ \hspace{0.3cm} Slip (49) \hspace{0.3cm} }} &
    \multicolumn{2}{c|}{\rule{0pt}{2.6ex}\textit{ Utensil (141) } } &
    \multicolumn{2}{c}{\rule{0pt}{2.6ex}\textit{ Measurement (25) }} \\
    \cline{2-9}
    & \multicolumn{1}{c}{\scriptsize{\textbf{S} (8)}} & \multicolumn{1}{c|}{\scriptsize{\textbf{U} (12)}} &
      \multicolumn{1}{c}{\scriptsize{\textbf{S} (7)}} & \multicolumn{1}{c|}{\scriptsize{\textbf{U} (24)}} &
      \multicolumn{1}{c}{\scriptsize{\textbf{S} (34)}} & \multicolumn{1}{c|}{\scriptsize{\textbf{U} (24)}} &
      \multicolumn{1}{c}{\scriptsize{\textbf{S} (5)}} & \multicolumn{1}{c}{\scriptsize{\textbf{U}} (0)} \\
    \hline
    ZS &  0.0 & 0.0 & 0.0 & 0.0 & 0.0 & 0.0 & 0.0 &  - \\
    ZS + \textit{R }  & \textbf{50.0} & 0.0 & 14.3 & 4.2 & 69.7 & 50.0 & \textbf{80.0} &  - \\
    SFT            &   0.0 & 0.0 & 0.0 & 0.0 & 2.9 & 8.3 & 0.0 & - \\
    SFT + \textit{R } & 12.5 & 0.0 & 14.3 & 0.0 & 23.5 & 16.7 & \underline{60.0} &  - \\
    SFTe            &  0.0 & 0.0 & 0.0 & 0.0 & 5.9 & 0.0 & 0.0 & - \\
    GRPO           & 37.5 & 0.0 & \textbf{57.1} & \textbf{95.8} & \underline{88.2} & \underline{70.8} & \underline{60.0} & - \\
    \hline
    \rowcolor{orange!30!white} \textbf{Ours} & \textbf{50.0} & 0.0 & \textbf{57.1} & \underline{91.7} & \textbf{91.2} & \textbf{83.3} & \textbf{80.0} & - \\
  \bottomline
\end{tabular}
}
\caption{\textbf{Per-Error Recall for EP-VQA seen and unseen test splits.} We show the number of samples for each mistake type in parentheses for seen (S) and unseen (U) splits. The number of training error samples is shown next to the name of the corresponding mistake category. Since the unseen split does not contain measurement mistakes, we report “–” for those entries. Our method consistently scores among the best methods in mistake recall for both seen and
unseen instructions. }
\label{tab:egoper_per_mistake}
\end{table*}

\section{Question and Prompt Templates}
\label{sec:supp_quest_prompts}

\subsection{Question Templates}

As discussed in the main paper, during training, we diversify the prompts to avoid overfitting to a specific wording or question format. In particular, during training, we employ the following question formulations. \\

\noindent For \textbf{EP-VQA}:
\begin{itemize}
    \item \raggedright \texttt{For the given video and an instruction text `<TEXT>', does the video show any mistakes being made while performing the step? The following mistake types are possible: Technique error, Slip error, Utensil error and Measurement error.}
    \vspace{0.2cm}
    \item \texttt{Given the video clip and a cooking step `<TEXT>', are there any errors being made while performing the step? The following mistake types are possible: Technique error, Slip error, Utensil error and Measurement error.}
    \vspace{0.2cm}
    \item \texttt{Given the video clip and a cooking step `<TEXT>', is there any mismatch between the video clip and the text instruction? The following mistake types are possible: Technique error, Slip error, Utensil error and Measurement error.}
    \vspace{0.2cm}
    \item \texttt{For the given video clip and the cooking step `<TEXT>', are there any errors in the text step execution? The following mistake types are possible: Technique error, Slip error, Utensil error and Measurement error.}
\end{itemize}

\noindent For \textbf{CC-VQA}:
\begin{itemize}
    \item \raggedright \texttt{For the given video and an instruction text `<TEXT>', does the video show any mistakes being made while performing the step? The following mistake types are possible: Technique error, Preparation error, Timing error, Temperature error and Measurement error.}
    \vspace{0.2cm}
    \item \texttt{Given the video clip and a cooking step `<TEXT>', are there any errors being made while performing the step? The following mistake types are possible: Technique error, Preparation error, Timing error, Temperature error and Measurement error.}
    \vspace{0.2cm}
    \item \texttt{Given the video clip and a cooking step `<TEXT>', is there any mismatch between the video clip and the text instruction? The following mistake types are possible: Technique error, Preparation error, Timing error, Temperature error and Measurement error.}
    \vspace{0.2cm}
    \item \texttt{For the given video clip and the cooking step `<TEXT>', are there any errors in the text step execution? The following mistake types are possible: Technique error, Preparation error, Timing error, Temperature error and Measurement error.} 
    \vspace{0.2cm}
\end{itemize}

\noindent During evaluation, we rely on a single fixed question formulation. \\

\noindent For \textbf{EP-VQA}:
\begin{itemize}
    \item \raggedright \texttt{Given the video clip and a cooking step `<TEXT>', is there any mismatch between the video clip and the text instruction? The following mistake types are possible: Technique error, Slip error, Utensil error and Measurement error.}
\end{itemize}

\noindent For \textbf{CC-VQA}:
\begin{itemize}
    \item \raggedright \texttt{Given the video clip and a cooking step `<TEXT>', is there any mismatch between the video clip and the text instruction? The following mistake types are possible: Technique error, Preparation error, Timing error, Temperature error and Measurement error.}
    \vspace{0.2cm}
\end{itemize}

\noindent In these prompts, \texttt{`<TEXT>'} is replaced with the textual description of the correct execution of the step.

\subsection{Prompt Templates}
For the task of MD-VQA, the questions, as described above, are concatenated together with one of the two types of prompts. The \textbf{base prompt}, which does not enforce explicit reasoning, is used for the Zero-Shot (ZS), Supervised Fine-Tuning (SFT) (without reasoning), and Supervised Fine-Tuning with explanation (SFTe) baselines. The \textbf{reasoning prompt}, on the other hand, encourages the model to perform step-by-step reasoning, as in~\cite{shao2024deepseekmath}, and is used for the ZS+R, SFT+R, GRPO baselines and our proposed method.

\paragraph{Base prompt}\mbox{}\\
\raggedright \texttt{"Answer YES if there is a mistake, otherwise answer NO.\textbackslash nProvide your final answer between <answer> </answer> tags."}\\

\paragraph{Reasoning Prompt}\mbox{}\\
\raggedright \texttt{"Answer YES if there is a mistake, otherwise answer NO.\textbackslash nPlease think about this question as if you were a human pondering deeply.\textbackslash nEngage in an internal dialogue using expressions such as `let me think', `wait', `Hmm', `oh, I see', `let's break it down', etc, or other natural language thought expressions.\textbackslash nIt's encouraged to include self-reflection or verification in the reasoning process.\textbackslash nProvide your detailed reasoning between the <think> </think> tags, and then give your final answer between the <answer> </answer> tags."} \\

\section{Data Cleaning and Processing}
\label{sec:supp_data_cleaning}
\subsection{Mapping EP-VQA}
In the original EgoPER dataset~\cite{lee2024egoper}, steps performed with mistakes do not include the corresponding \textit{counter-descriptions} that specify how the same steps should be executed correctly. To address this limitation, we manually create additional annotations that describe the correct execution of each mistake step. The full mapping between mistake steps and their corresponding correct-execution descriptions is provided in Tab.~\ref{tab:newanns}.
In addition to that, some normal steps have an ambiguous description. We clarify the annotations for those steps, as shown in Tab.~\ref{tab:egoper_new_anns}. 

\subsection{CaptainCook4D Data Cleaning}

The annotations of CaptainCook4D contain a number of inconsistencies and artifacts. To make sure the baselines and our model use clean data, we perform a number of cleaning steps.
First, the original annotations are in the format ``\textless \textit{verb}\textgreater - \textless \textit{verb}\textgreater (...)", for example, ``\textit{cut-cut english muffin}", and we remove this duplication. Second, we correct the grammar wherever needed. (\eg \textit{Cut or tear 1 slices} to \textit{Cut or tear 1 slice}).
Third, we make sure that clips labeled as a mistake actually contain one. For instance, for the description \textit{Cut the English muffin into two pieces with a knife}, it would not be a mistake to cut the muffin into two uneven pieces, because it is not specified in the instruction nor inherently wrong. In this case, we then specify \textit{Cut the English muffin into two even pieces with a knife}. 
For every mistake clip and description, we additionally provide \textit{why} there is a mistake to train SFTe and as a possible avenue for future works. For example, \textit{Pour 2 eggs into the ramekin cup} becomes \textit{Pour 2 eggs into the ramekin cup instead of 1 egg}.

\begin{table*}[tbp]
    \centering
    \resizebox{0.9\linewidth}{!}{%
    \begin{tabular}{c|c}
        \topline
        \textbf{Original description of step with mistake} & \textbf{Correct execution of the step} \\
        \hline
        \rowcolor{black!15!white} \multicolumn{2}{c}{\textit{Pinwheels}} \\
        \hline
        Drop tortilla on floor & Place tortilla on cutting board \\
        Place tortilla on table & Place tortilla on cutting board \\
        Slice using knife & Slice using floss \\
        Fold tortilla & Roll tortilla \\
        Use spoon to scoop nut butter & Use knife to scoop nut butter \\
        Use spoon to spread butter onto tortilla & Use knife to spread butter onto tortilla \\
        Clean spoon & Clean knife \\
        Use spoon to scoop jelly & Use knife to scoop jelly \\
        Use spoon to spread jelly on nut butter & Use knife to spread jelly on nut butter \\

        \hline
        \rowcolor{black!15!white} \multicolumn{2}{c}{\textit{Quesadilla}} \\
        \hline
        Drop tortilla & Place tortilla on cutting board \\
        Place tortilla on table & Place tortilla on cutting board \\
        Fold tortilla into quarter-circle & Fold tortilla in half \\
        Place tortilla wedges into bowl & Place tortilla wedges on plate \\
        Rip tortilla by hands & Slice using knife \\

        \hline
        \rowcolor{black!15!white} \multicolumn{2}{c}{\textit{Tea}} \\
        \hline
        Directly pour water to kettle & Measure 12 ounces of cold water \\
        Pour water into another mug & Pour water from kettle into mug with teabag \\
        Stir using knife & Stir using spoon \\
        Add sugar to mug & Add honey to mug \\
        Drop tea bag & Place tea bag in mug \\

        \hline
        \rowcolor{black!15!white} \multicolumn{2}{c}{\textit{Oatmeal}} \\
        \hline
        Directly pour quick oats into bowl & Measure 4 Tablespoons of quick-cook oats \\
        Add bananas to another empty bowl & Put bananas on oats \\
        Pour sugar instead of honey & Drizzle honey in bowl \\
        Add water to a different bowl & Pour water to the bowl with oats \\
        Stir using knife & Stir using spoon \\

        \hline
        \rowcolor{black!15!white} \multicolumn{2}{c}{\textit{Coffee}} \\
        \hline
        Hold cup of coffee with filter cone in front of you & Hold cup of coffee in front of you \\
        Squeeze paper filter into cone without folding & Place paper filter in dripper and spread it into cone \\
        Measure incorrect weight of coffee beans & Weigh 25 grams of coffee beans \\
        Pour water without circular motion & Slowly pour the rest of water in circular motion \\
        Measure incorrect amount of water & Measure 12 ounces of cold water \\
        Pour too much water to overflow mug & Slowly pour the rest of water in circular motion \\
        Tear paper filter & Fold paper filter in half to create semi-circle \\
        Spill out coffee grounds & Transfer grounds to filter cone \\
        Spill out coffee beans & Put coffee beans in coffee grinder \\
        Grind coffee beans without enough time & Grind coffee for 20 seconds \\
        \bottomline
    \end{tabular}
    }
    \caption{\textbf{New annotations for EP-VQA.} The left column shows the original descriptions of the steps executed with mistakes; the right column contains the new annotations representing the correct execution of the corresponding step.}
    \label{tab:newanns}
\end{table*}

\begin{table}[tbp]
    \centering
    \resizebox{0.6\linewidth}{!}{%
        \begin{tabular}{c|c}
        \topline
        Original & Updated \\
        \hline
        \rowcolor{black!15!white} \multicolumn{2}{c}{Pinwheels} \\
        \hline
        Spread butter onto tortilla & Use knife to spread butter onto tortilla \\
        Spread jelly on nut butter & Use knife to spread jelly on nut butter \\

        \hline
        \rowcolor{black!15!white} \multicolumn{2}{c}{Quesadilla} \\
        \hline
        Fold tortilla & Fold tortilla in half \\
        
        \hline
        \rowcolor{black!15!white} \multicolumn{2}{c}{Tea} \\
        \hline
        Pour water into mug & Pour water into mug with teabag \\

        \hline
        \rowcolor{black!15!white} \multicolumn{2}{c}{Oatmeal} \\
        \hline
        Put bananas & Put bananas on oats \\
        Pour water to bowl & Pour water to bowl with oats \\
        \bottomline
        \end{tabular}
    }
    \caption{\textbf{Updated annotations for ambiguous step descriptions on EP-VQA.}}
    \label{tab:egoper_new_anns}
\end{table}

\section{Additional Qualitative Results}
\label{sec:supp_qual}
\begin{figure*}[h!]
    \centering
    \includegraphics[width=\linewidth]{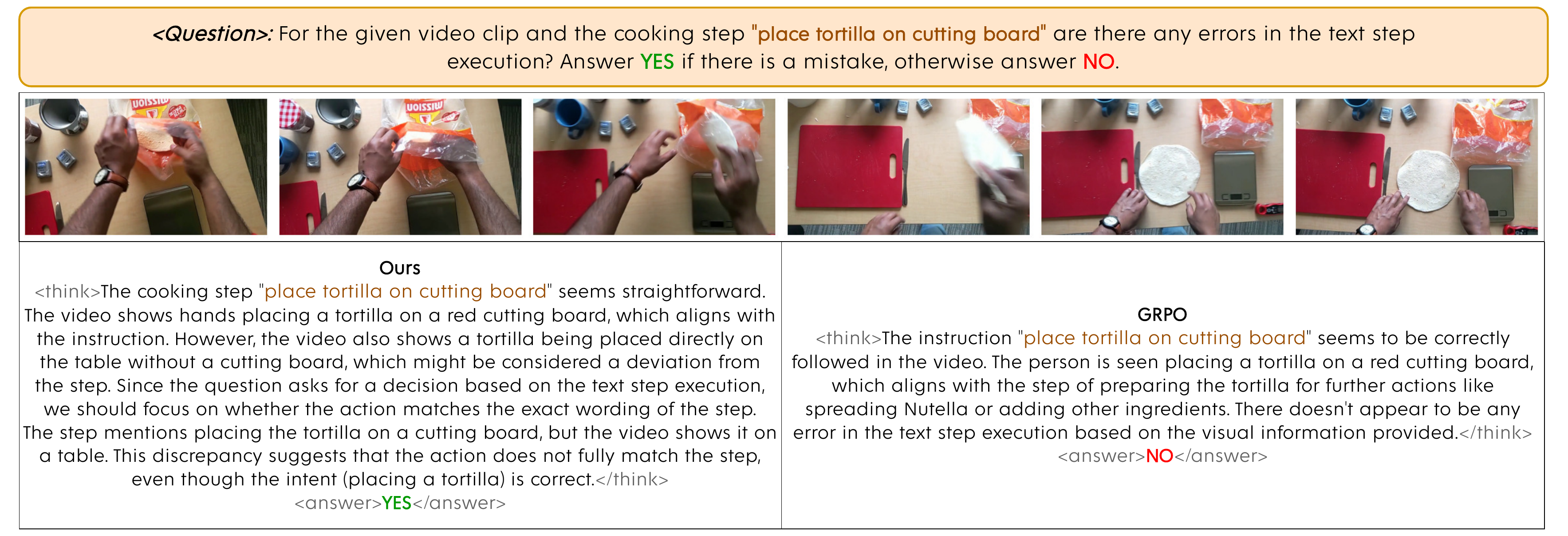}
    \caption{\textbf{EP-VQA test seen} split for the Pinwheels activity. The step requires placing the tortilla on a cutting board, but the participant places it directly on the table, leading to a mistake.}
    \label{fig:egoper-seen-qual2}
\end{figure*}

\begin{figure*}[h!]
    \centering
    \includegraphics[width=\linewidth]{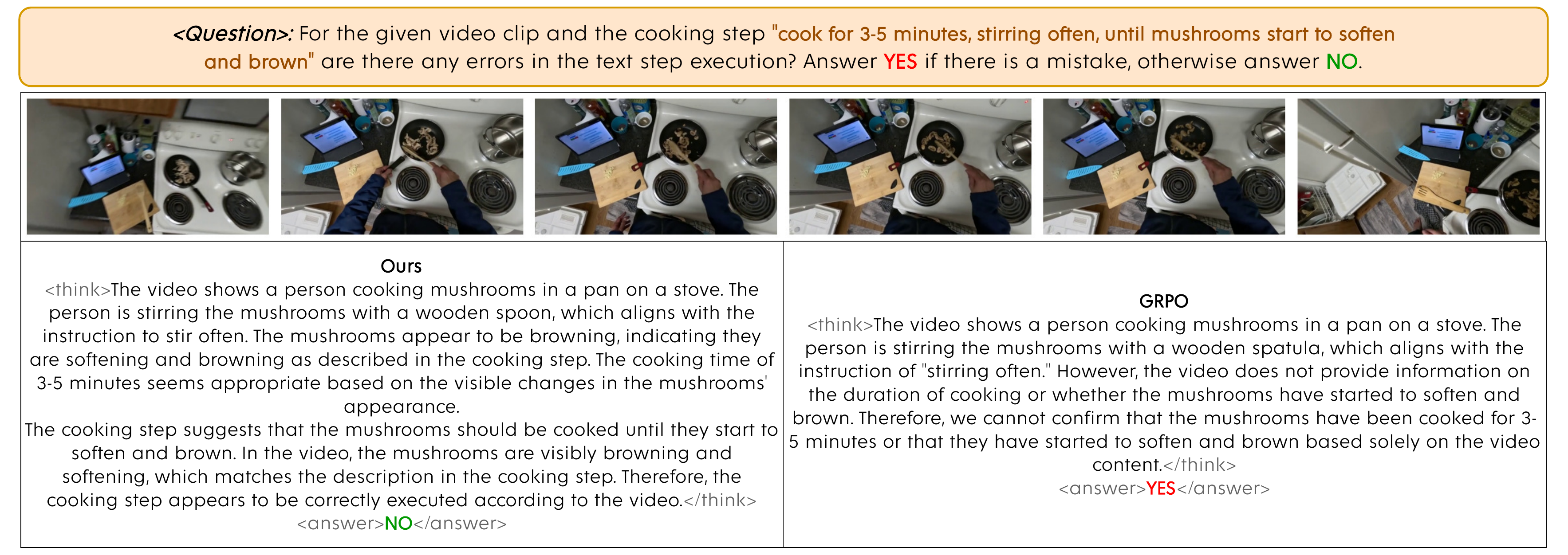}
    \caption{\textbf{CC-VQA test seen} split. The step requires the participant to cook the mushrooms for a duration of 3–5 minutes, with the goal of softening and browning them. Standard GRPO focuses too heavily on the exact cooking time rather than the actual objective, whereas our method captures the intended outcome.}
    \label{fig:cc4d-seen-qual}
\end{figure*}

\begin{figure*}[h!]
    \centering
    \includegraphics[width=\linewidth]{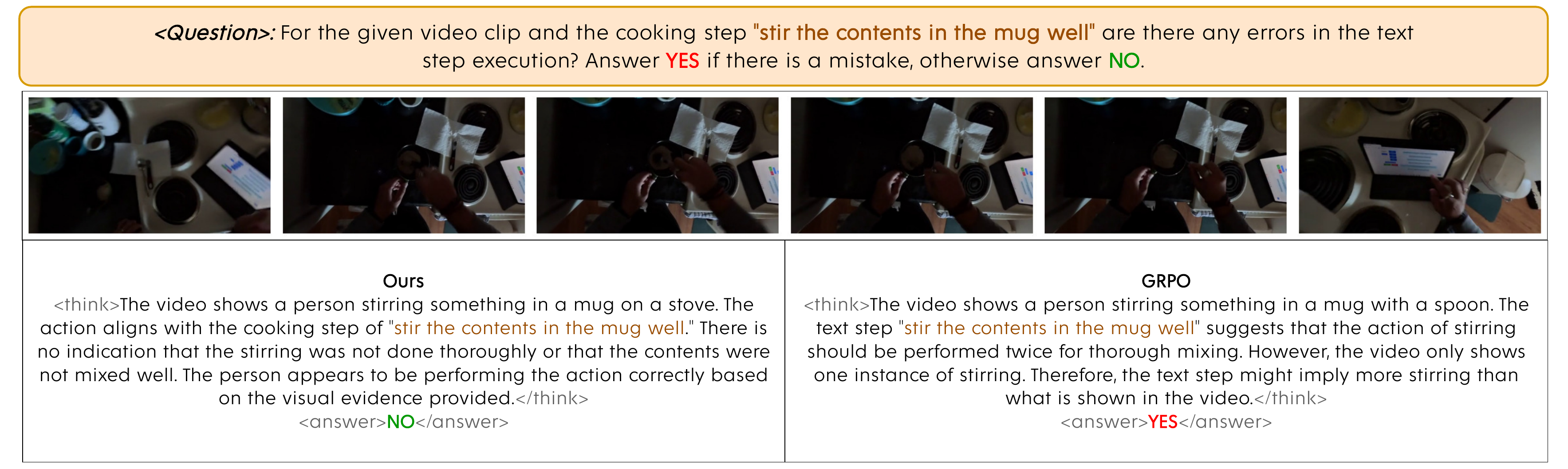}
    \caption{\textbf{CC-VQA test unseen} split. The video shows the participant stirring the contents of the mug multiple times, indicating a correct execution of the step “stir the contents in the mug well”. However, standard GRPO misclassifies the action as a mistake, failing to account for the multiple stirrings performed.}
    \label{fig:cc4d-unseen-qual}
\end{figure*}

We present additional qualitative examples from both the seen and unseen splits of CC-VQA and EP-VQA. Fig.~\ref{fig:egoper-seen-qual2} shows an example from the EP-VQA seen test split, Fig.~\ref{fig:cc4d-seen-qual} from the CC-VQA seen test split, and Fig.~\ref{fig:cc4d-unseen-qual} from the CC-VQA unseen test split.

Fig.~\ref{fig:egoper-seen-qual2} depicts a case in which the participant places a tortilla on the table rather than on the cutting board as instructed. Although the cutting board is visible in the scene and the initial part of the instruction is executed correctly, our method accurately detects that the final placement does not satisfy the step specification.

Fig.~\ref{fig:cc4d-seen-qual} illustrates a task where the goal is to cook the mushrooms until they are soft and brown, for an approximate duration of 3–5 minutes. As the time is approximate, the visual texture of the food is more important. Our method captures the change in color and consistency, even though it does not have access to the full duration of the process, whereas GRPO focuses excessively on the time duration.

The final CC-VQA example, shown in Fig.~\ref{fig:cc4d-unseen-qual}, depicts the task of stirring the contents of a mug “well”. Although the term “well” can be ambiguous, both our method and GRPO correctly reason that stirring multiple times is sufficient to achieve a well-mixed result. However, even when the stirring is repeated multiple times in the video, GRPO fails to recognize it and consequently misclassifies the action as incorrect.

%  (video \textit{pinwheels\_u1\_a1\_error\_010\_1})

%  (video \textit{20\_14\_20\_213\_0\_360p})

%  (video \textit{3\_2\_3\_38\_0\_360p})

\section{Further ablations}

\subsection{Results with newer model}

\begin{table}[t]
    \centering
    \renewcommand{\arraystretch}{1.15}
    \setlength{\tabcolsep}{5pt}
    \begin{tabular}{l|ccc|ccc}
        \topline
        & \multicolumn{6}{c}{\textbf{EP-VQA}} \\
        \cline{2-7}
        \textbf{Method} & \multicolumn{3}{c|}{\textbf{Seen}} & \multicolumn{3}{c}{\textbf{Unseen}} \\
        \cline{2-7}
        & \textit{Rec} & \textit{Prec} & \textbf{\textit{F1}} & \textit{Rec} & \textit{Prec} & \textbf{\textit{F1}} \\
        \hline
        ZS & 11.1 & 17.6 & 13.6 & 16.7 & 45.5 & 24.4 \\
        ZS + R & 68.5 & 21.8 & 33.0 & 48.3 & 36.7 & 41.7 \\
        GRPO & \textbf{81.5} & \textbf{25.6} & \textbf{38.9} & 66.7 & 34.5 & 45.5 \\
        \textbf{Ours} & 79.2 & 23.3 & 36.1 & \textbf{71.7} & \textbf{41.3} & \textbf{52.4} \\
        \hline
        \bottomline
    \end{tabular}
\caption{\textbf{Results with Qwen3 on EP-VQA.} Our method also works with a more recent model.}
    \label{tab:qwen3}
    % %\vspace{-0.3cm}
    %\vspace{-1pt}
\end{table}

To show that our method also works with a newer model, we evaluate Qwen3-VL-8B-Instruct on EP-VQA in Table~\ref{tab:qwen3}. The results show that Qwen3 is a stronger zero-shot baseline than Qwen2.5, both with and without reasoning.
However, the task remains challenging, and our method improves over the two Qwen3 baselines, suggesting that the proposed training strategy is still useful even when starting from a more recent backbone. Compared to standard GRPO, our method slightly underperforms on the seen split, but again achieves a large improvement on the unseen split.

\subsection{Comparison to non-VLM methods}

As the mistakes and tasks in the test data are not the same as in the training data, we can not compare our approach to task-specific error detection models, as these can not handle the open-set scenario. 
The only other feasible class of models that can be evaluated zero-shot in this setting are contrastive CLIP-style models. We evaluate CLIP (ViT large patch14, 0.4B parameters) by comparing the text-video similarity of ``a person correctly performing the cooking step: \textless description\textgreater" to that of  
``a person incorrectly performing the cooking step: \textless description\textgreater" and selecting the highest one.

In Tab.~\ref{tab:clip}, we compare CLIP on EP-VQA to our approach. Our method outperforms CLIP by a large margin, indicating that the reasoning capabilities of the VLM are crucial for the complex task of detecting subtle mistakes, thereby also justifying the use of large-scale models. 

\begin{table}[t]
    \centering
    \renewcommand{\arraystretch}{1.15}
    \setlength{\tabcolsep}{5pt}
    % \resizebox{\columnwidth}{!}{
    \begin{tabular}{l|ccc|ccc}
        \topline
        & \multicolumn{6}{c}{\textbf{EP-VQA}} \\
        \cline{2-7}
        \textbf{Method} & \multicolumn{3}{c|}{\textbf{Seen}} & \multicolumn{3}{c}{\textbf{Unseen}} \\
        \cline{2-7}
        & \textit{Rec} & \textit{Prec} & \textbf{\textit{F1}} & \textit{Rec} & \textit{Prec} & \textbf{\textit{F1}} \\
        \hline
        CLIP & 46.3 & 11.6 & 18.6 & 8.3 & 9.8 & 9.0 \\
        \rowcolor{orange!30!white} \textbf{Ours} & {\textbf{79.6}} & {\textbf{40.6}} & {\textbf{53.8}} & {\textbf{70.0}} & {\textbf{36.5}} & {\textbf{48.0}} \\
        \bottomline
    \end{tabular}
    % }
\caption{\textbf{Comparison to CLIP model on EP-VQA.} Our method greatly outperforms CLIP, indicating that reasoning capabilities are necessary for strong performance.}
    \label{tab:clip}
    % %\vspace{-0.3cm}
    %\vspace{-1pt}
\end{table}

\subsection{Disentangling contributions}

We perform an additional ablation study to isolate the benefits of chain-of-thought (CoT) and the opposite reward for GRPO post-training.
As shown in Table~\ref{tab:ablcot}, GRPO + Opposite Reward without CoT predicts a mistake for almost all samples, shown by the 100\% recall for the error class with very low precision. 
As such, CoT plays an important role in preventing degenerate solutions when using the Opposite Reward. 
Furthermore, using CoT with GRPO but without the opposite reward underperforms our proposed method. This shows that the gains come from the mistake-specific supervision rather than the reinforcement learning itself. Finally, excluding both CoT and the opposite reward similarly underperforms.
Overall, both CoT and Opposite Reward are essential to perform well on both seen and unseen instructions.

\begin{table}[t]
    \centering
    \renewcommand{\arraystretch}{1.15}
    \setlength{\tabcolsep}{5pt}
    \begin{tabular}{l|ccc|ccc}
        \topline
        & \multicolumn{6}{c}{\textbf{EP-VQA}} \\
        \cline{2-7}
        \textbf{Method} & \multicolumn{3}{c|}{\textbf{Seen}} & \multicolumn{3}{c}{\textbf{Unseen}} \\
        \cline{2-7}
        & \textit{Rec} & \textit{Prec} & \textbf{\textit{F1}} & \textit{Rec} & \textit{Prec} & \textbf{\textit{F1}} \\
        \hline        
        GRPO + Opp w/o CoT & 100.0 & 8.3 & 15.3 & 100.0 & 10.1 & 18.4 \\
        GRPO + CoT w/o Opp & {74.1} & 39.2 & {51.3} & {66.7} & {31.7} & {43.0} \\         
        GRPO w/o CoT \& Opp & 77.8 & 29.8 & 43.1 & 68.3 & 36.0 & 47.1 \\
        \rowcolor{orange!30!white} \textbf{Ours} & {79.6} & {40.6} & {53.8} & {70.0} & {36.5} & {48.0} \\
        \bottomline
    \end{tabular}
\caption{\textbf{Additional experiments on EP-VQA.}  }
    \label{tab:ablcot}
    % %\vspace{-0.3cm}
    %\vspace{-1pt}
\end{table}

\subsection{Importance of textual context}
\label{sec:supp_add_abl}

We investigate whether conditioning mistake detection on the textual description of the correct step is necessary for our proposed protocol, or whether providing only visual information is sufficient for models to determine the correctness of a given video clip. To this end, we evaluate the (ZS + \textit{R}) baseline without providing the textual description of the correct step and only ask whether the video contains a mistake. 

As shown in Tab.~\ref{tab:bench_macroF1_seen_unseen}, the step descriptions are necessary for accurate mistake detection. In many cases, the mistake is defined as a deviation from the instruction, for example, using an incorrect amount of an ingredient, rather than an action that is inherently wrong in all situations. Therefore, without the reference point provided by the correct instructional step description, mistake detection is often ill-defined and cannot be addressed effectively.

\begin{table}[h!]
\centering
\resizebox{0.6\textwidth}{!}{
    \begin{tabular}{l|cc|c|cc|c}
    \topline
    & \multicolumn{3}{c|}{\textbf{Seen}} 
    & \multicolumn{3}{c}{ \textbf{Unseen}} \\
    \hline
    \textbf{Method} & \textit{ Prec } & \textit{ Rec } & \textbf{\textit{ F1 } } 
    & \textit{ Prec } & \textit{ Rec } & \textbf{\textit{ F1 } } \\
    \hline
    \rowcolor{orange!30!white} ZS + \textit{R} w. instr   &  \textbf{12.5} & \textbf{60.4} & \textbf{20.7} & \textbf{12.3} & \textbf{21.7} & \textbf{15.7} \\
    ZS + \textit{R} w/o instr   & 5.3 & 7.7 & 6.3 & 9.3 & 6.8 & 7.8  \\
    \bottomline
    \end{tabular}
    \vspace{-0.5cm}
}
\caption{\textbf{Ablating instruction-aware mistake detection.} We analyze the effect of removing textual step instructions on performance for seen and unseen test splits on EP-VQA using (ZS+\textit{R}) baseline. Detecting mistakes without context is ill-defined.}
\label{tab:bench_macroF1_seen_unseen}
\end{table}

% \bibliography{egbib}
% \end{document}

\end{document}